\documentclass[10pt,journal,compsoc]{IEEEtran}

\ifCLASSOPTIONcompsoc
  \usepackage[nocompress]{cite}
\else
  \usepackage{cite}
\fi

\usepackage[cmex10]{amsmath}
\usepackage{amssymb}
\usepackage{amsthm}

\usepackage{url}
\usepackage{ragged2e}

\usepackage{algorithm}
\usepackage{algpseudocode}
\usepackage{times}

\usepackage{xspace}
\usepackage{enumitem}
\usepackage{threeparttable}
\usepackage{amsmath}
\usepackage{amssymb}
\usepackage{algorithm}
\usepackage{caption}
\usepackage{algpseudocode}
\usepackage{epsfig}
\usepackage{graphicx}
\usepackage{bm}
\usepackage{multirow}
\usepackage{subfigure}
\usepackage{verbatim}
\usepackage{soul, xcolor}
\usepackage{color}

\begin{document}
\title{Patient Flow Prediction via Discriminative Learning of Mutually-Correcting Processes}

\author{Hongteng~Xu,~ %~\IEEEmembership{Fellow,~OSA,}
        Weichang~Wu,~ %~\IEEEmembership{Member,~IEEE,}
        Shamim Nemati,~ %~\IEEEmembership{Fellow,~OSA,}
        and~Hongyuan~Zha, \\%~\IEEEmembership{Fellow,~OSA,}
\IEEEcompsocitemizethanks{
\IEEEcompsocthanksitem H. Xu is with the School of Electrical and Computer Engineering, Georgia Institute of Technology, Atlanta, GA, 30332.
E-mail: hxu42@gatech.edu
\IEEEcompsocthanksitem W. Wu is with Shanghai Key Lab of Digital Media Transmission and Processing, Department of Electronic Engineering, Shanghai Jiao Tong University, Shanghai, China, 200240.
E-mail: blade091@sjtu.edu.cn
\IEEEcompsocthanksitem S. Nemati is with the School of Medicine, Emory University, Atlanta, GA, 30322.
E-mail: shamim.nemati@emory.edu
\IEEEcompsocthanksitem H. Zha are with the College of Computing, Georgia Institute of Technology, Atlanta, GA, 30332. E-mail: zha@cc.gatech.edu}
}

\markboth{IEEE Transactions on Knowledge and Data Engineering,~Vol.~XX, No.~X, September~201X}%
{Xu \MakeLowercase{\textit{et al.}}: Patient Flow Prediction via Discriminative Learning of Mutually-Correcting Processes}

\IEEEtitleabstractindextext{
\begin{abstract}\justifying{
Over the past decade the rate of care unit (CU) use in the United States has been increasing. 
With an aging population and ever-growing demand for medical care, effective management of patients' transitions among different care facilities will prove indispensible for shortening the length of hospital stays, improving patient outcomes, allocating critical care resources, and reducing preventable re-admissions. 
In this paper, we focus on an important problem of predicting the so-called ``patient flow'' from longitudinal electronic health records (EHRs), which has not been explored via existing machine learning techniques.
By treating a sequence of transition events as a point process, we develop a novel framework for modeling patient flow through various CUs and jointly predicting patients' destination CUs and duration days. 
Instead of learning a generative point process model via maximum likelihood estimation, we propose a novel discriminative learning algorithm aiming at  improving the prediction of transition events in the case of sparse data. 
By parameterizing the proposed model as a mutually-correcting process, we formulate the estimation problem via generalized linear models, which lends itself to efficient learning based on alternating direction method of multipliers (ADMM). 
Furthermore, we achieve simultaneous feature selection and learning by adding a group-lasso regularizer to the ADMM algorithm. 
Additionally, for suppressing the negative influence of data imbalance on the learning of model, we synthesize auxiliary training data for the classes with extremely few samples, and improve the robustness of our learning method accordingly. 
Testing on real-world data, we show that our method obtains superior performance in terms of accuracy of predicting the destination CU transition and duration of each CU occupancy.}
\end{abstract}

\begin{IEEEkeywords}
Patient flow; mutually-correcting process; discriminative learning; logistic regression; group lasso; imbalanced data.
\end{IEEEkeywords}}

\maketitle

\IEEEdisplaynontitleabstractindextext

\IEEEpeerreviewmaketitle

\IEEEraisesectionheading{\section{Introduction}\label{sec:introduction}}
\IEEEPARstart{R}{ecent} reports have highlighted an increasing demand for care units in the United States due to an improved life expectancy and a larger aging population~\cite{institute2006future}. 
Patient management and reducing waiting time, particularly in the Emergency Department (ED)~\cite{trzeciak2003emergency,olshaker2009managing} and intensive care unit (ICU)~\cite{pascual2014there,knight2015complications}, is crucially important to improving quality of care, outcomes, and the overall patient satisfaction. 
The so-called practice of ``patient boarding'' refers to temporarily keeping critically-ill patients in their existing hospital location, such as the emergency department or the post anesthesia unit, while awaiting available CU bed~\cite{pascual2014there}, which may result in suboptimal care, and increase both length of stay (LOS) and hospital mortality~\cite{chalfin2007impact,ziser2002postanaesthesia}. 
System-level management of medical resources becomes even more critical for large numbers of critically-ill patients in the case of disasters and pandemics~\cite{dichter2014system}. 

\begin{figure*}[!t]
\centering
\subfigure[An example of patient flow.]{
\includegraphics[width=0.38\linewidth]{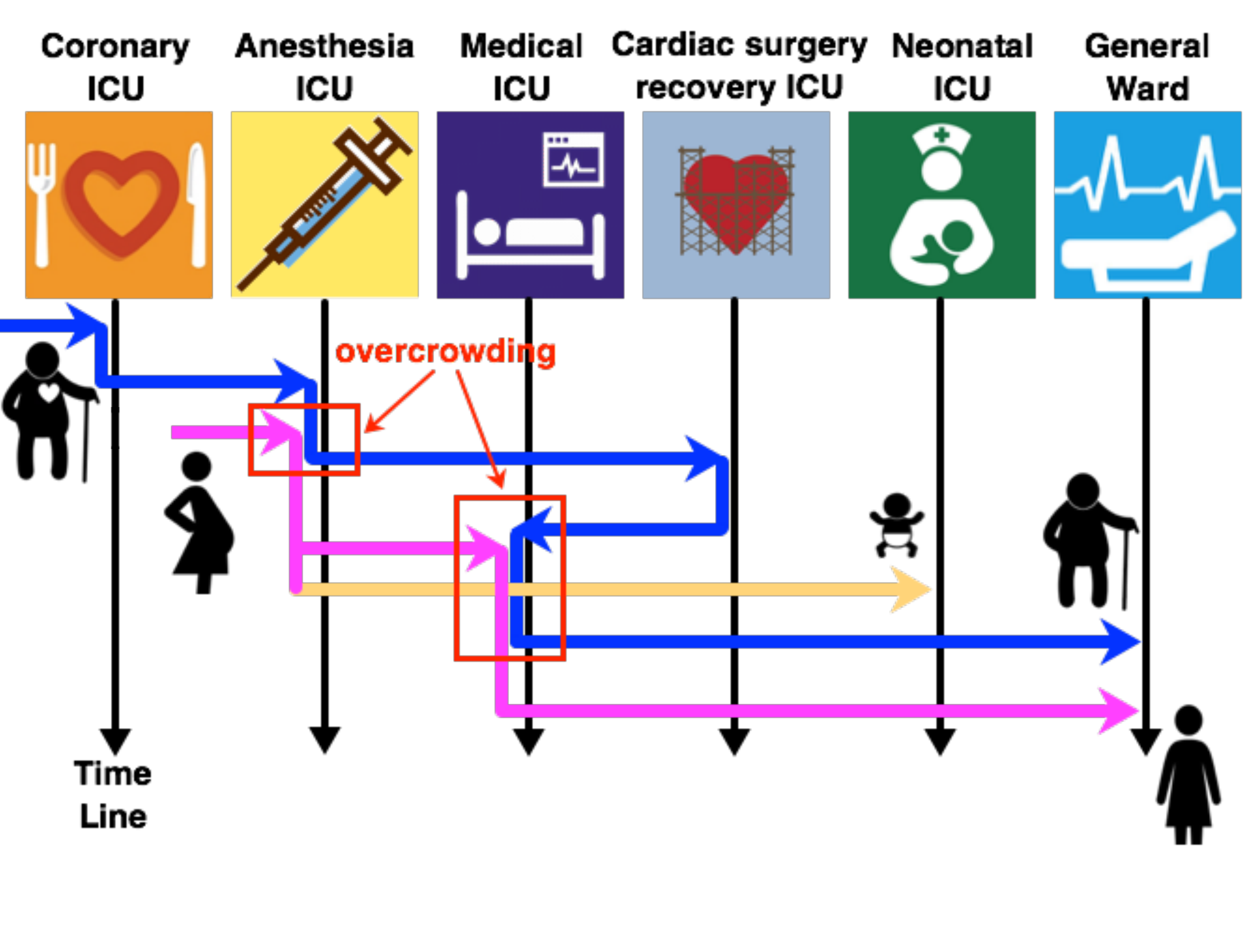}\label{Fig1a}
}
\subfigure[The principle of proposed method.]{
\includegraphics[width=0.56\linewidth]{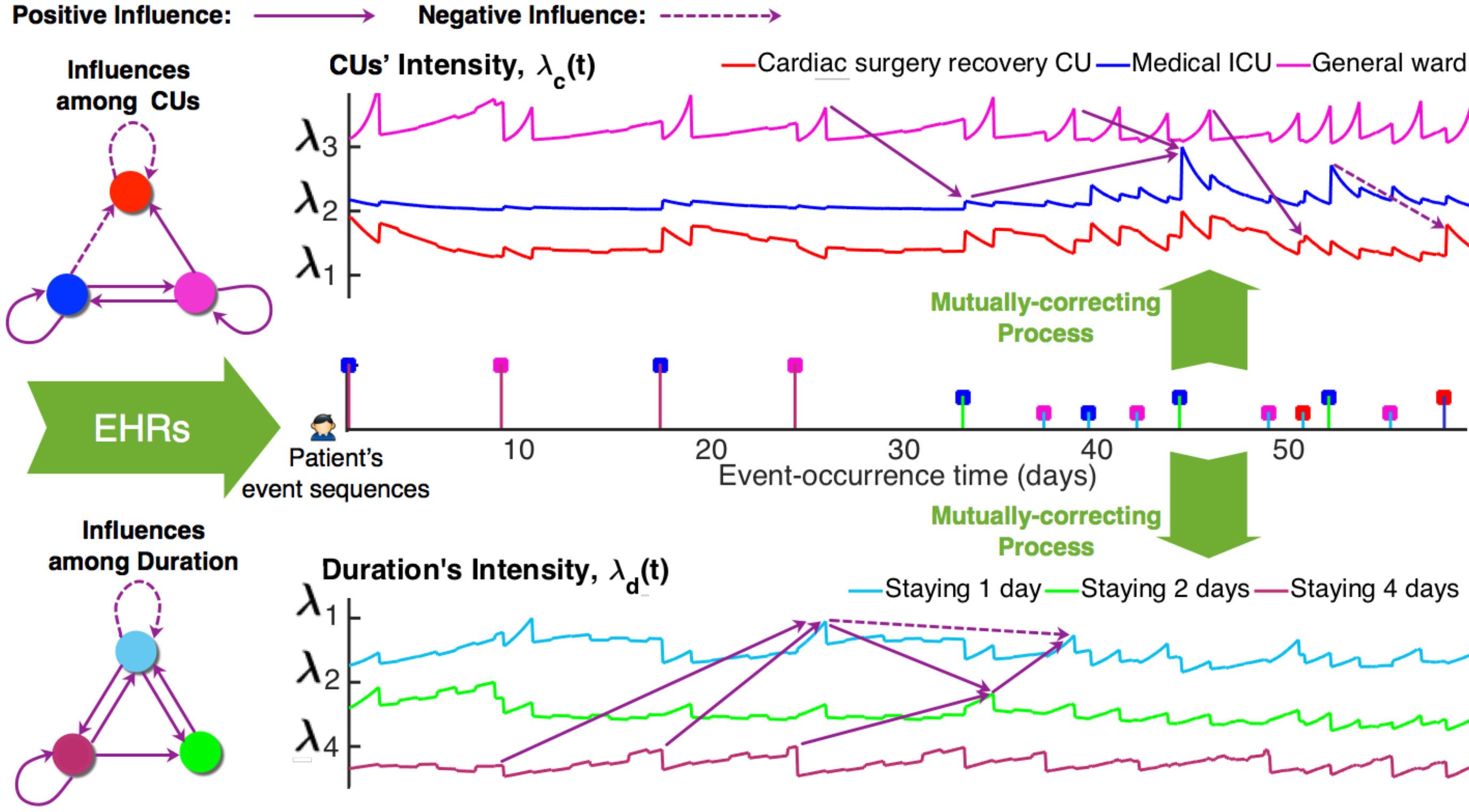}\label{Fig1b}
}
\caption{
(a) The transition process of an old male patient having coronary heart disease may include the Coronary Care Unit for preoperative tests, the Anesthesia Services for cardiac surgery, the Cardiac Surgery Recovery unit, and finally the Medical ICU and general ward for nursing.
During this period, the transition process of a pregnant woman having a premature baby may include the Anesthesia Services for a Caesarean section surgery, the Medical Care Unit for the mother, and the Neonatal Care Unit for the baby. 
There is a partial overlap between the need of the elderly patient and that of the pregnant woman for anesthesia services and within the Medical ICU, which may cause scheduling conflicts and may require advanced planning and scheduling to reduce waiting times. 
(b) The transition process of patient is represented via two event sequences of destination CUs and duration days respectively.
Along the time line, the color dots indicate various CUs and the color lines with various length indicate the durations (in units of days). 
Applying our mutually-correcting process model, the conditional intensity functions for CUs and durations are proposed to capture the positive and negative influences among unit types and durations, respectively.}
\end{figure*}

Such an urgent requirement gives rise to an important problem of predicting the transition processes of patients, known as the ``patient flow'' (see Fig.~\ref{Fig1a}), which has not been explored via existing machine learning techniques.
The patient flow includes patients' duration time within each care unit and transition probability among different units, and is determined by a number of factors including patient's underlying condition and clinical state, disease progression, and availability of care team and care resources. 
With the advent of comprehensive electronic health records (EHRs) and real-time streaming analytics~\cite{blount2010real}, much of these factors can be captured and utilized to jointly model flow of patients within many care units. 
Therefore, the problem we aim to address in this work involves predicting patients' destination CUs and durations simultaneously based on their medical records and continuously-documented clinical status. 
Solving this problem may enable early planning and optimization of hospital resources.

However, predicting patient flow is a difficult task due to a number of factors, e.g., the collection and the storage of a huge amount of data, the lack of a systematic approach to resource management, etc. 
Additionally, from the viewpoint of machine learning, the main challenges include:

\textbf{Time-sensitivity.} The prediction of patient flow is a time-sensitive learning task, which requires us to both predict the destination care unit of a patient (i.e., the transition) and the dwell time within that care unit (i.e., the duration). 

\textbf{Feature selection.} The patient flow can be viewed as a time-varying transition process in continuous time, which is influenced by many medical factors, e.g., patients' health profiles, diagnoses, medications, nursing, etc. 
However, the relationships between these factors and the transition process are not fully explored and their importance for predicting patient flow is unknown. 
Although modern EHRs may include complete or partial information pertaining to most of these factors, taking advantage of EHRs involves feature selection and fusion, all of which are highly dependent on the model used to describe the patient flow process. 

\textbf{Data sparsity and case imbalance.} Because most patients more often stay in general wards than  transfer to other CUs (or moved around within the same CU),  models and learning algorithms may suffer from sparse and imbalanced data --- the general ward appears in most of transition processes while a certain CU may only appear in very few of them.

Considering the challenges above, we need a predictive model that jointly captures the transitions and durations in patient flow. 
Moreover, the model should consider all influential factors and be robust to data sparsity and imbalance. 
To the best of our knowledge, no existing work has been proposed to deal with such a challenging situation.  
For achieving this aim, in this paper we propose a novel and efficient method that utilizes both time-invariant and time-varying features from patients' EHRs, to predict the patient flow, as depicted in Fig.~\ref{Fig1b}. 

Based on the unique characteristics of patient flow, we consider the transitions among the care units and the dwell time within each care unit as two separate events, which are jointly modeled via a novel parametric point process model called mutually-correcting process. 
Applying our mutually-correcting process model with the EHR-based features, the instantaneous rate of patient being transferred to a given patient care unit and that of staying certain days in the unit are captured via two parameterized conditional intensity functions. 
Compared with traditional models, such as discrete Markov chain~\cite{cole2005multistate}, vector auto-regressive model~\cite{han2013transition} and hidden Markov model~\cite{cooper2004analysis}, which can only deal with time-invariant transition process formulated as discrete time series, our point process model is able to describe time-varying transition processes in continuous time. 
\textcolor{black}{Compared with other continuous model, such as the continuous-time Markov chain~\cite{anderson2012continuous}, our model captures the mutually-correcting patterns among states over time using all historical data, which does not need to set the order of model in advance. 
In other words, our model is more robust to sparse data and model misspecification.} 

Besides proposing a mutually-correcting process to model the patient flow, another technical contribution of our work 
is the development of a methodology for learning a parametric point process model in a discriminative way. 
Specifically, traditional generative point processes model the joint distribution of events in continuous time and parameters are learned via the maximum likelihood estimation. 
In this work, however, we focus on learning the conditional distribution of transition and that of duration given historical events. 
We analyze the relationship between the conditional distribution and the conditional intensity function, showing that by using the proposed mutually-correcting process, we can formulate the learning problem as learning a multinomial logistic regression model that greatly simplifies the learning task.  
\textcolor{black}{Thousands of factors, i.e., diagnoses, treatments and medications, are generated and accompany the patient flow, while only few of them are highly influential. 
Moreover, these key factors generally have influences on both patients' destination CUs and duration at the same time.
Therefore, feature selection is introduced into the framework of our learning algorithm. 
Specifically, we formulate these factors as high-dimensional features, which are shared via the logistic regressor for predicting destination CUs and that for predicting duration. 
The parameters of these two models are associated with these features and learned jointly. 
We treat each dimension of feature (i.e., a factor influencing patient flow) as a ``group'' and regularize the parameters via $l_{1,2}$-norm. 
It guarantees the group sparsity of parameters so that only the parameters corresponding to the features of important dimensions are non-zero and shared via both models.} 
Leveraging the alternating direction method of multipliers (ADMM)~\cite{gabay1976dual} with group-lasso~\cite{simon2013sparse}, we propose an efficient algorithm to learn the model. 

\textcolor{black}{For overcoming the data imbalance problem, we investigate several robust learning methods for imbalanced data and make comparisons for them. 
We focus on applying a pre-processing on our imbalanced patient flow data, which shows its superiority in our experiments: for the classes with extremely few samples, we synthesize some auxiliary samples from original ones to increase the number of training samples. 
Taking original samples and auxiliary ones as training samples, we can improve the robustness of our learning method greatly and obtain better performance in the testing phase.}

In summary, the contributions of our method include: 
1) We propose a flexible mutually-correcting process model to capture the properties of patient flow.  
2) We propose a discriminative algorithm to learn point processes. 
In certain cases, e.g., the proposed mutually-correcting processes, the algorithm can be implemented via logistic regression. 
3) Combining group-lasso with ADMM, we achieve feature selection and learning model jointly in the training phase. 
4) The influence of data imbalance is considered, and a preprocessing step is applied to synthesize auxiliary data for training. The preprocessing helps us to improve the robustness of learning algorithm. 

Our method can be viewed as a point process-based interpretation of multinomial logistic regression model for continuous-time transition processes. 
We test our method on real-world patient flow data set and compare it with several alternative methods on the prediction accuracy and robustness to imbalanced data. 
Additionally, we analyze the functions of various parameters and investigate their impacts on our learning algorithm. 
We demonstrate the robustness of our method to those parameters. 
Multiple metrics are applied to evaluate the performance of various methods, including the prediction accuracy of the destination CUs and duration days, and the relative simulation error of the patient flow. 
Experimental results demonstrate that our method significantly outperforms its competitors, especially in predicting those unique transitions and usages of CUs. 

%The rest of the paper is organized as follows.
%We first formally define the problem and analyze the properties of data in section~\ref{sec3}. 
%The proposed model and the learning algorithm are given in section~\ref{sec:model}. 
%Section~\ref{sec:exp} shows the experimental results and demonstrate the superiority of our method.
%The related work is given in section~\ref{sec2:relatedwork} and section~\ref{sec:con} concludes the paper.

\section{Background and Data Analysis}\label{sec3}
\subsection{Notations and Problem Statement}
Suppose that we have $U$ patients in a hospital having $C$ CU departments.
For each patient $u$, $u=1,...,U$, her transition process among CUs is represented via an event sequence in continuous time, denoted as $s^u=\{(c_i^u, d_i^u, t_i^u)\}_{i=1}^{N^u}$. 
Here, $t_i^u \in (0, T^u]$ is the time when a transition event happened, $T^u$ is the length of observation time window, $c_i^u \in\mathcal{C}$, $\mathcal{C}=\{1,...,C\}$, is the destination CU of the transition, $d_i^u\in\mathcal{D}$, $\mathcal{D}=\{1,...,D\}$, is the dwell time (measured by the number of duration days) of the patient in the previous CU (i.e., the $c_{i-1}^{u}$-th CU) before the transition, and $N^u$ is the number of transitions\footnote{When $i=1$, we do not consider the duration and set $d_i^u=\mbox{NULL}$.}. 
The set of historical transitions before time $t$ is denoted as $\mathcal{H}_t^u=\{(c_i^u, d_i^u, t_i^u)| t_i^u<t\}$.

Each event $(c,d,t)$, which means that a patient stays in a CU for $d$ days before transferred to the $c$-th CU, is always accompanied by a series of medical services. 
According to the EHRs of patients, we classify various medical services into three categories: treatment, medication and nursing. 
The treatment contains $M_{treat}$ items, including various medical tests, surgeries and therapies. 
The medication contains $M_{med}$ items, including various medicines and their various usage methods. 
The nursing contains $M_{nurse}$ items, including various nursing programs and records of patients' liquid inputs and outputs. 
We can extract binary feature vectors for patient $u$ from her EHRs, denoted as $\bm{f}_i^u\in\{0,1\}^{M_{treat}+M_{med}+M_{nurse}}$, $i=1,...,N_u$.
Here $\bm{f}_{i}^u$ is a binary vector corresponding to the EHR of patient $u$ when staying in the $c_{i}^u$-th CU, in which the elements corresponding to received services are $1$'s. 
It is the concatenation of three binary vectors corresponding to the three categories above. 
Besides the time-varying features mentioned above, a patient's EHR also contains $M_{p}$ time-invariant features, including personal health profile like gender, age, chronic diseases, and diagnoses\footnote{In our data set the diagnose is time-invariant because the patient flow for each patient is collected after a single diagnose.}. 
Similarly, we can extract a binary feature vector for the patient, denoted as $\bm{f}_{0}^u\in\{0,1\}^{M_p}$. 

The event sequence of patient can be modeled using point process methodology~\cite{daley2007introduction}. 
Specifically, we capture the temporal dynamics of event sequences via the conditional intensity function defined as follows:
\begin{eqnarray}\label{intensity}
\begin{aligned}
\lambda(t)dt=\mathbb{E}(dN(t)|\mathcal{H}_t),
\end{aligned}
\end{eqnarray}
where $N(t)$ is the number of events occurred in time range $(-\infty, t]$, $\mathcal{H}_t$ contains  historical events before time $t$, and $\mathbb{E}(dN(t)|\mathcal{H}_t)$ is the expectation of the number of events happening in the interval $(t,t+dt]$ given historical observations $\mathcal{H}_t$.
The conditional intensity function in Eq.~(\ref{intensity}) represents the expected instantaneous rate of future events at time $t$.
Based on conditional intensity function, the conditional probability that an event happens at time $t$ given historical record is computed as
\begin{eqnarray}\label{probability}
\begin{aligned}
p(t|\mathcal{H}_t)=\lambda(t)\exp\left(-\int_{t_I}^{t}\lambda(s)ds\right).
\end{aligned}
\end{eqnarray}
Here $t_I$ is the time stamp of the last event before time $t$. 

\textbf{Problem statement.} For each patient $u$, given historical record $\mathcal{H}_{t_{i-1}^u}^u$ and EHR features $\{\bm{f}_{0}^u, \bm{f}_{1}^u,...,\bm{f}_{i-1}^u\}$, we aim to predict the destination CU of the next transition (i.e., $c_{i}^u$) and the duration before the transition (i.e., $d_{i}^u$).

%\vspace{3pt}
\subsection{Data and Basic Statistics}
We focus on the real-world data from MIMIC II database~\cite{goldberger2000physiobank}, from which $30,685$ patients staying in CUs are selected for training and testing. 
The CUs are categorized into $C=8$ departments, including the Coronary care unit (\textbf{CCU}), the Anesthesia care unit (\textbf{ACU}), the Fetal ICU (\textbf{FICU}), the Cardiac surgery recovery unit (\textbf{CSRU}), the Medical ICU (\textbf{MICU}), the Trauma Surgical ICU (\textbf{TSICU}), the Neonatal ICU (\textbf{NICU}), and the general ward (\textbf{GW}). 
According to the EHRs of the patients, the number of treatment items is $M_{treat}=5,627$, the number of medication items is $M_{med}=405$, the number of nursing items is $M_{nurse}=6,808$, and the number of time-invariant features is $M_p=4,832$. 

The data is representative, which reflects the following natures of patient flow. 
For each department, the number of patients ever staying in it and the number of transitions directing to it are shown in Table~\ref{tab:stats}.
We can find that the data for various departments is imbalanced. 
On the one hand, most of the patients and transitions concentrate on certain CUs, e.g., GW, CCU, and CSRU, etc.
On the other hand, few patients and transitions involve ACU and TSICU. 
The average duration days for each department is also listed. 
Except for NICU, the average dwell time of other department is within one week. 
To simplify our treatment, we categorize the duration times into $D=8$ time intervals, include $1$ day, $2$ days, ...., $7$ days and more than $1$ week.

\begin{figure}[h!]
\centering
\includegraphics[width=1\linewidth]{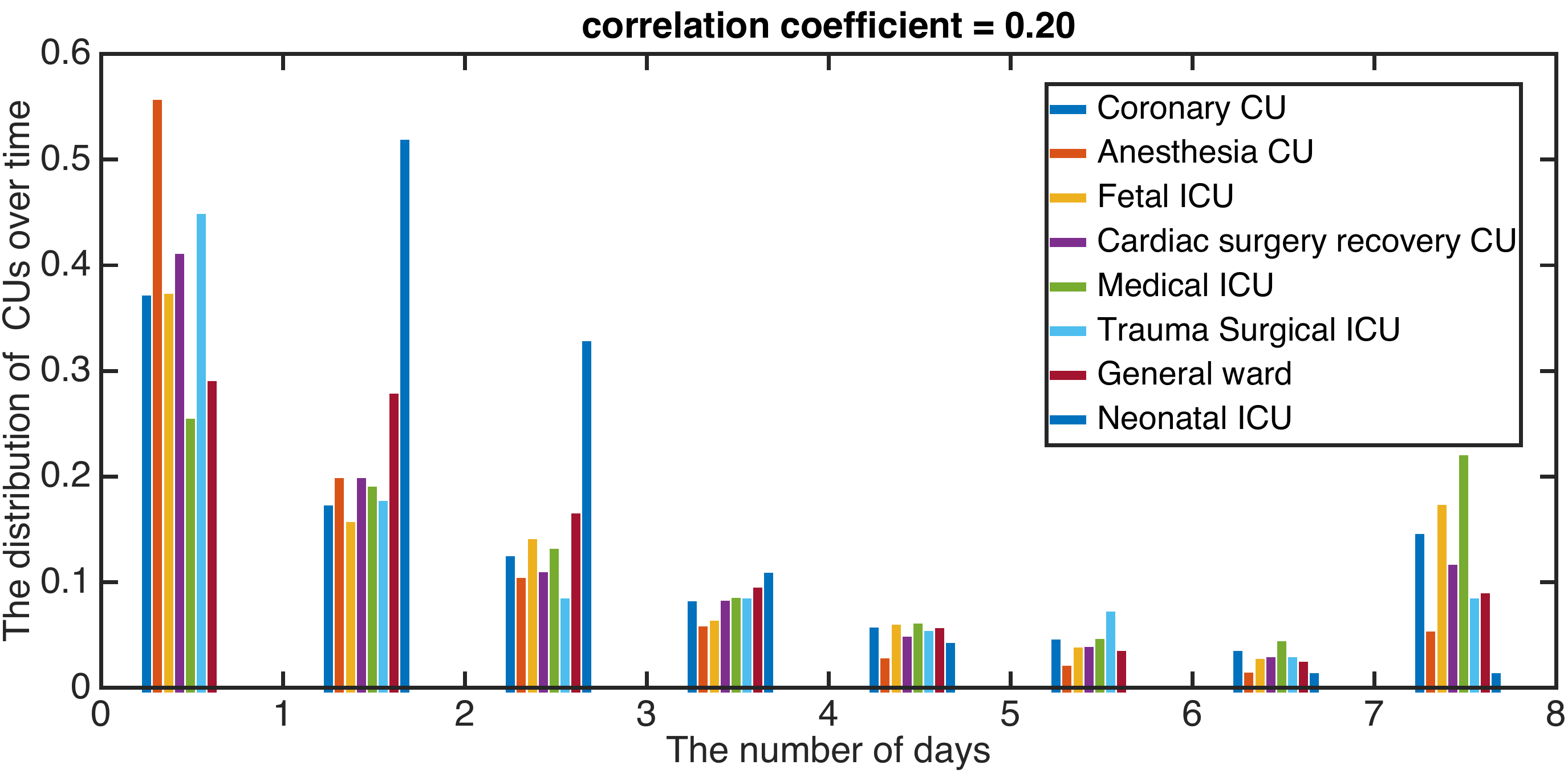}
\caption{
Each bin in the interval $[d-1, d]$ can be viewed as the probability that patients stay in the corresponding CU $d$ days after transferring. 
The correlation coefficient between the transition destination and duration is given on the top.}\label{Fig2}
\end{figure}

Interestingly we also observed that the transitions and the durations are weakly correlated with each other. 
The correlation coefficient between the transition and the duration is about $0.2$. 
Fig.~\ref{Fig2} further gives the normalized histograms of various CUs w.r.t. the categories of duration days. 
We can find that in each category of duration, the frequency of occurrence for various CUs generally do not have large variance. 
It should be noted that the nature of the weak correlation between the transition and the duration is important for us to simplify our model, which will be shown in the following section.

Table~\ref{tab:stats2} gives the proportions of nonzero elements in different feature domains w.r.t various CUs. 
Specifically, we count the number of nonzero elements in different feature domains for each CU and normalize the counts.  
The proportions reflect the importance of feature domains. 
We can find that patient's profile, treatment, and nursing are relatively important for all CUs, which contain most of nonzero features. 
On the contrary, the proportion of nonzero features from medication is relatively low. 
For TSICU and GW, most of nonzero features concentrate in the domain of treatment.

\begin{table}[ht!]
  \centering
  \small\caption{Number of patients and transitions, and average durations (days) in each CU.}\label{tab:stats}
  \begin{threeparttable}[c]
      \begin{tabular}{
        @{\hspace{2pt}}l@{\hspace{2pt}}
        @{\hspace{2pt}}r@{\hspace{2pt}}
        @{\hspace{2pt}}r@{\hspace{2pt}}
        @{\hspace{2pt}}r@{\hspace{2pt}}
        @{\hspace{2pt}}r@{\hspace{2pt}}
        @{\hspace{2pt}}r@{\hspace{2pt}}
        @{\hspace{2pt}}r@{\hspace{2pt}}
        @{\hspace{2pt}}r@{\hspace{2pt}}
        @{\hspace{2pt}}r@{\hspace{2pt}}
        }
        \hline\hline
Depts.         &CCU   &ACU  &FICU  &CSRU   &MICU  &TSICU   &NICU   &GW\\ \hline
\# patients    &6,259 &559  &3,254 &9,490  &7,245 &1,552 &7,458 &23,748\\
\# trans.      &7,030 &631  &3,525 &10,679 &8,903 &1,628 &7,657 &28,118\\ 
durations      & 3.32 &2.38 &4.46  &3.96   &3.83  &3.21  &9.01  &4.15\\
        \hline\hline
      \end{tabular}
   \end{threeparttable}
\end{table}

\begin{table}[ht!]
  \centering
  \small\caption{The proportions of nonzero elements in different feature domains w.r.t various CUs.}\label{tab:stats2}
  \begin{threeparttable}[c]
      \begin{tabular}{
        @{\hspace{2pt}}l@{\hspace{2pt}}
        @{\hspace{2pt}}r@{\hspace{2pt}}
        @{\hspace{2pt}}r@{\hspace{2pt}}
        @{\hspace{2pt}}r@{\hspace{2pt}}
        @{\hspace{2pt}}r@{\hspace{2pt}}
        @{\hspace{2pt}}r@{\hspace{2pt}}
        @{\hspace{2pt}}r@{\hspace{2pt}}
        @{\hspace{2pt}}r@{\hspace{2pt}}
        @{\hspace{2pt}}r@{\hspace{2pt}}
        }
        \hline\hline
Depts.    &CCU   &ACU   &FICU  &CSRU   &MICU  &TSICU   &NICU   &GW\\ \hline
Profile   &0.347 &0.512  &0.347  &0.330  &0.513  &0.001 &0.640 &0.001\\
Treatment &0.505 &0.354  &0.505  &0.562  &0.342  &0.995 &0.241 &0.996\\
Nursing   &0.117 &0.112  &0.120  &0.085  &0.121  &0.002 &0.100 &0.001\\
Medication&0.031 &0.022  &0.028  &0.023  &0.024  &0.002 &0.019 &0.002\\
        \hline\hline
      \end{tabular}
   \end{threeparttable}
\end{table}

%\vspace{-7pt}
\section{Proposed Method}\label{sec:model}
In this section, we take advantage of the properties of patient flow and propose a mutually-correcting point process to describe the transitions among CUs and the durations in them respectively. 
The proposed model can be viewed as a specialization of a generalized parametric point process model. 
\textcolor{black}{It has higher capability and can represent more complicated temporal dynamics of event sequences than existing popular point processes, e.g., modulated Poisson process~\cite{cole2005multistate}, Hawkes process~\cite{luo2015multi} and self-correcting process~\cite{xu2015trailer}.} 
A discriminative learning algorithm for the point process model is proposed, which combines the alternating direction method of multipliers (ADMM) and the group-lasso. 
Both the feature selection problem and the imbalance of data are considered in our learning algorithm.
Finally, a pre-processing method for training samples is proposed to handle the data imbalance problem.

\subsection{Mutually-correcting Process Model}
As aforementioned, patient flow is a time-varying transition process in continuous time. 
It generally has two important properties. 
Again, take the patient flow in Fig.~\ref{Fig1a} as an example: 

\textbf{High correlation between EHRs and patient flow.} 
\textcolor{black}{A typical EHR consists of a patient's profile (i.e., gender, age), her diagnose of certain diseases (i.e., ICD code), and her treatment process, e.g., medications, nursing information, the transitions and durations in various care units. 
It reflects the patient's status and contains very useful information for predicting patient flow. 
Recall the previous cases shown in Fig.~\ref{Fig1a}. 
For a man having coronary heart disease, the probability staying in the Coronary care unit is relatively high, while the probability staying in the Neonatal ICU is zero. 
On the contrary, for a premature baby, the probability staying in the Neonatal ICU is high while the probability staying in the Coronary care unit is very low.
In more general cases, most of patients whose treatments involve surgeries are likely to have transitions among the Anesthesia care unit, the surgery recovery unit, and the general ward. 
Similarly, the duration of a patient in a CU is also dependent on her health record. 
The patients having chronic diseases may spend a lot of time at the general ward. 
The patients after surgeries may stay at the surgery recovery units for varying time according to their feedback of treatments and recovery. 
In summary, the patient flow is highly correlated with their EHRs.
The patients' EHR can help us to predict what types of CUs they need and how long will they stay at different CUs.} 

\textbf{Mutually-correcting across CUs.} Staying in the Coronary care unit is likely to increase the probability transferring to the Cardiac surgery recovery unit while suppress the probability transferring to the Neonatal ICU. 
It reflects that the duration of previous CU has a positive or negative influence on the transitions to following CUs, which is called mutually-correcting in our work.

Therefore, both the transitions among CUs and the durations in different CUs contain mutually-correcting patterns, which are highly dependent on EHR-based features. 
Additionally, taking the weak correlation between the transition and the duration (Fig.~\ref{Fig2}) into consideration, we propose a new point process model called mutually-correcting process to model the transitions and the durations respectively. 
Specifically, given the event sequence $s^u=\{c_i^u, d_i^u, t_i^u\}_{i=1}^{N_u}$ of patient $u$, we decouple the event $(c,d)$ into two independent events $c$ and $d$, which correspond to two counting processes $\{N_{c}^u(t)\}_{c=1}^{C}$ and $\{N_d^u(t)\}_{d=1}^{D}$.  
Here $N_{c}^u(t)$ is the number of events that transferring patient $u$ to the $c$-th CU after time $t$, while $N_d^u(t)$ is the number of events that staying in a CU $d$ days after time $t$.  
We propose a generalized parametric model for the conditional intensity functions of these two counting processes as follows,\footnote{It should be noted that Eq.~(\ref{lambdaC}) can be further generalized by replacing  $\bm{\alpha}g(t)$, $\bm{\beta}h(t, t_i^u)$ with functional $\bm{\alpha}(t)$, $\bm{\beta}(t)$. 
Then, the model becomes nonparametric, which is out-of-range in this paper.}
\begin{eqnarray}\label{lambdaC}
\begin{aligned}
\lambda_{c}^u(t)=f( \bm{\alpha}_{c}^{\top}\bm{f}_0^u g(t) - \bm{\beta}_{c}^{\top}\sideset{}{_{i: t_{i}^{u}<t}}\sum\bm{f}_{i}^u h(t, t_i^u)),\\
%\end{aligned}
%\end{eqnarray}
%\begin{eqnarray}\label{lambdaD}
%\begin{aligned}
\lambda_{d}^u(t)=f( \bm{\alpha}_{d}^{\top}\bm{f}_0^u g(t) - \bm{\beta}_{d}^{\top}\sideset{}{_{i: t_{i}^{u}<t}}\sum\bm{f}_{i}^u h(t, t_i^u)).
\end{aligned}
\end{eqnarray}
$\lambda_c^u(t)$ represents the instantaneous  rate of the event transferring patient $u$ to the $c$-th CU at time $t$, while $\lambda_d^u(t)$ represents the instantaneous  rate of the event staying in a CU $d$ days. 
Here $\{\bm{f}_0^u, \bm{f}_i^u\}$ are time-invariant and time-varying features defined in Section~\ref{sec3}. 
The term $\bm{\alpha}^{\top}\bm{f}_{0}^u g(t)$ represents the temporal influence of time-invariant feature of the patient on event. 
The term $\bm{\beta}^{\top}\sum_{i: t_i^u<t}\bm{f}_i^u$ represents the temporal influences of historical transitions $\mathcal{H}_t^u$ on event. 
Here $f(\cdot)$, $g(\cdot)$ and $h(\cdot, \cdot)$ are predefined time functions, which describes the increase or the decay of influences over time.

\begin{figure}[!ht]
\centering
\includegraphics[width=1\linewidth]{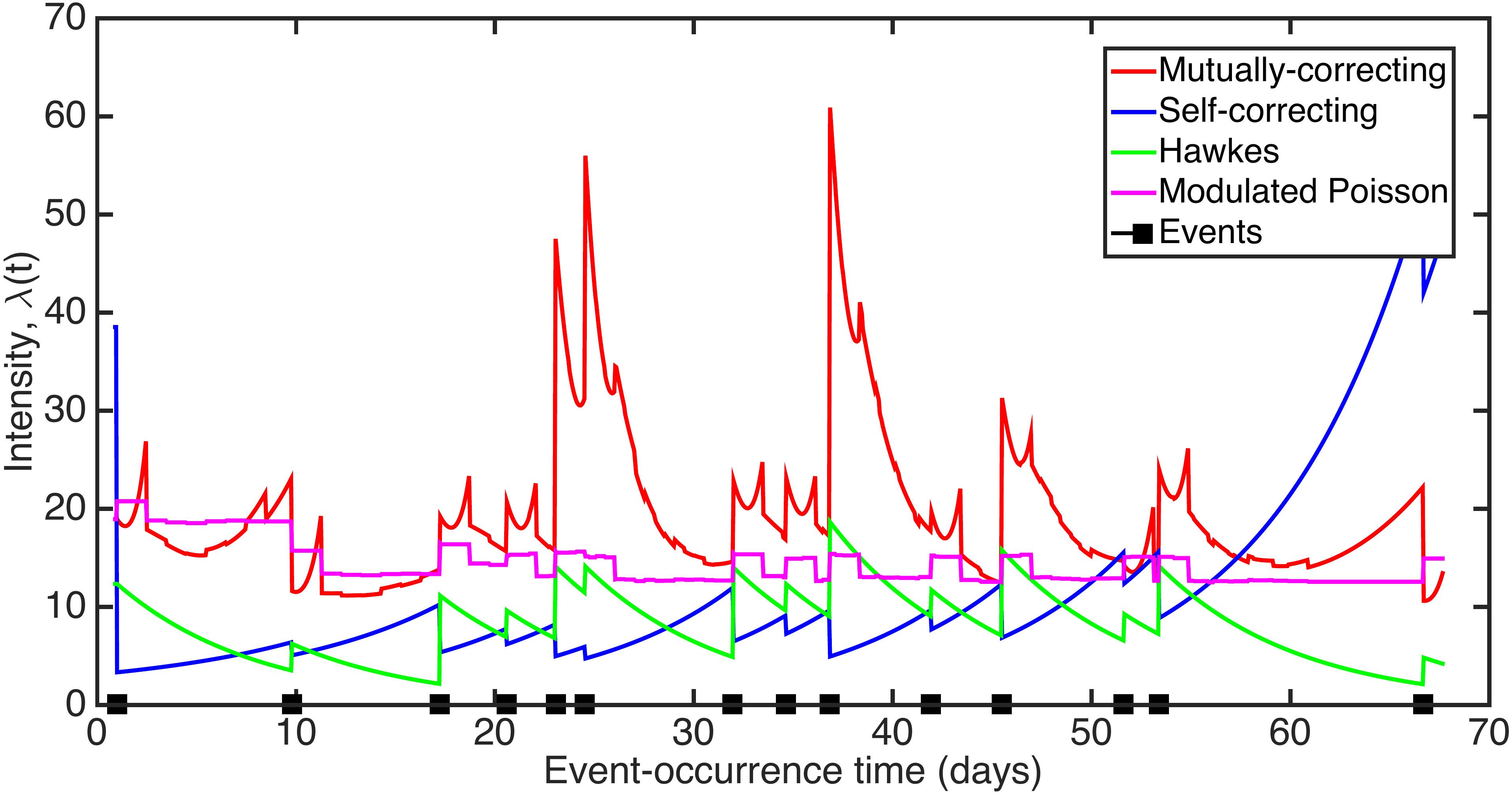}
\caption{
Comparison on conditional intensity function for various point processes. An event sequence is given and the conditional intensity functions of various point processes are shown.}\label{FigIntensity}
\end{figure}

\begin{table}[ht!]
  \centering
  \small\caption{Comparison of various parametric point processes.}\label{tab:PP}
  \begin{threeparttable}[c]%\begin{small}
      \begin{tabular}{
        @{\hspace{2pt}}l@{\hspace{2pt}}
        @{\hspace{2pt}}c@{\hspace{2pt}}
        @{\hspace{2pt}}c@{\hspace{2pt}}
        @{\hspace{2pt}}c@{\hspace{2pt}}
        @{\hspace{2pt}}c@{\hspace{2pt}}
        }
        \hline\hline
        Model  &$f(x)$  &$g(t)$  &$h(t,t')$  &Constraints  \\
        \hline
        Modulated Poisson process    &$x$    &$1$    &$1$ &$\bm{\beta}\leq \bm{0}\leq \bm{\alpha}$    \\
        Hawkes process                      &$x$    &$1$    &$e^{-w(t-t')}$    &$\bm{\beta}\leq \bm{0}\leq\bm{\alpha}$    \\
        Self-correcting process           &$e^x$   &$t$     &$1$  &$\bm{\alpha},\bm{\beta}\geq\bm{0}$\\
        \textbf{Mutually-correcting process}   &$e^x$ &$t-t_{I}$            &$e^{-\frac{(t-t')^2}{\sigma^2}}$    &---    \\
        \hline\hline
      \end{tabular}%\end{small}
      \begin{tablenotes}
 	 \begin{scriptsize}
         \item ``---'' means no constraints.\par
         \end{scriptsize}
  \end{tablenotes}
  \end{threeparttable}  
\end{table}

Eq.~(\ref{lambdaC}) provides a unified framework for many useful point processes, e.g., modulated Poisson processes~\cite{lloyd2014variational},  Hawkes processes~\cite{li2014learning,yan2015machine}, as Table~\ref{tab:PP} shows. 
In our mutually-correcting process model, we set $f(\cdot)=\exp(\cdot)$, $g(t)=t-t_{I}^u$, and $h(t,t')=\exp(-\frac{(t-t')^2}{\sigma^2})$, where $t_I^u$ is the time stamp of the last event before time $t$ for patient $u$. 
Our model extends traditional self-correcting process model~\cite{isham1979self} to multivariate case and further considers the temporal decay of influence from historical record. 
Compared with existing models, our model is more flexible. 
Firstly, different from self-correcting process, whose historical influence is time-invariant, i.e., $h(\cdot, \cdot)\equiv 1$, our model considers the time-varying historical influence as Hawkes process does. 
Secondly, for guaranteeing models to be physically-meaningful and stable, the self-correcting process requires all parameters $\bm{\alpha}=[\bm{\alpha}_1,...,\bm{\alpha}_C]$, $\bm{\beta}=[\bm{\beta}_1,...,\bm{\beta}_C]$ to be nonnegative while the modulated Poisson and Hawkes process require $\bm{\alpha}\geq 0$ and $\bm{\beta}\leq 0$. 
Our model, however, does not have such constraints. 
Such a relaxation increases the flexibility of our model and enhances the description power of conditional intensity function. 
Fig.~\ref{FigIntensity} shows that the dynamics of conditional intensity function for various point processes in 1-dimensional case. 
We can find that the conditional intensity function of modulated Poisson process is piecewise constant. 
A jump happens when a new event comes. 
However, the change of event's happening rate between adjacent events cannot be captured. 
Hawkes process and self-correcing process can only describe the change of event's happening rate via fixed pattern --- the conditional intensity always decreases for Hawkes process and increases for self-correcting process till new event comes. 
Our mutually-correcting process, however, is more flexible, which can capture both the increase and decrease of intensity function between adjacent events.

Obviously, the conditional intensity function of our mutually-correcting process model can be rewritten as 
\begin{eqnarray}\label{expintensity}
\begin{aligned}
\lambda_c^u(t)=\exp(\bm{\theta}_c^{\top}\bm{f}_t^u),\quad \lambda_d^u(t)=\exp(\bm{\theta}_d^{\top}\bm{f}_t^u).
\end{aligned}
\end{eqnarray}
$\bm{f}_t^u = [\bm{f}_{0}^{u\top}(t-t_I^u), (\sum_{t_i^u<t}\exp(-(t-t_i^u)^2/\sigma^2)\bm{f}_i^{u})^{\top}]^{\top}\in\mathbb{R}^M$, $\bm{\theta}_d=[\bm{\alpha}_d^{\top}, \bm{\beta}_d^{\top}]^{\top},~\bm{\theta}_c=[\bm{\alpha}_c^{\top}, \bm{\beta}_c^{\top}]^{\top}$, $M=M_{treat}+M_{med}+M_{nurse}+M_{p}$.
Such a simple representation inspires us to propose the following discriminative learning method for our model with the help of multinomial logistic regression.

\subsection{Discriminative Learning of Model}
Traditional learning methods for point processes are generative, which aim to estimate the joint probability of all events via a maximum likelihood estimator, i.e., $\max_{\bm{\Theta}}\prod_{u,i} p(c_i^u, d_i^u, t_i^u|\mathcal{H}_{t_i^u}^u)(1-P(T^u))$, where $p(c, d, t |\mathcal{H}_t^u)$ is the conditional probability of event $(c,d)$ given historical record $\mathcal{H}_t^u$, and $P(T^u)$ is the cumulative probability transferring before $T^u$. 
The parameters of the model is represented as a matrix $\bm{\Theta}=\{\bm{\theta}_c, \bm{\theta}_d\}_{c\in\mathcal{C},d\in\mathcal{D}}\in\mathbb{R}^{M\times (C+D)}$.  
However, the generative learning methods may lack discrimination power because it naturally cares more about the happening of the whole event sequence, than the classification or the prediction of individual events given historical record. 
The information of labels, e.g., the transition destination and the duration, is not fully used in the model. 
Additionally, the sparse and imbalanced data, e.g., the patient flow data we deal with, is insufficient for estimating the joint probability, so that the generative learning methods will be at high risk of over-fitting. 

According to the analysis above, we propose a discriminative learning method for our model. 
Recall the problem we have: given current time $t_{i-1}^{u}$ and historical record $\mathcal{H}_{t_{i-1}}^{u}$, we aim to maximize the probability that the patient $u$ stay in a CU $d_i^u$ days before being transferred to the $c_i^u$-th CU, i.e., $p(c_i^u,d_i^u| t_{i-1}^u, \mathcal{H}_{t_{i-1}}^u)$. 
Therefore, instead of estimating $p(c, d, t |\mathcal{H}_t^u)$ directly, we focus on the conditional probability $p(c,d| t, \mathcal{H}_t^u)$, which is the probability of event $(c,d)$ given current time $t$ and historical record. 
\textcolor{black}{As shown in Eq.~(\ref{lambdaC}), we decouple the event $(c,d)$ into two independent events $c$ and $d$, so we can specialize Eq.~(\ref{probability}) as}
\begin{eqnarray}\label{prob1}
\begin{aligned}
p(c, d, t | \mathcal{H}_t^u)=& \lambda_{c,d}^u(t)\exp\left(-\sum_{c'=1}^{C}\sum_{d'=1}^{D}\int_{t_I^{u}}^{t}\lambda_{c',d'}^u(s)ds\right)\\
=& \frac{\lambda_{c,d}^u(t)}{\sum\lambda_{c',d'}^u(t)}
\times\frac{\sum_{c',d'}\lambda_{c',d'}^u(t)}{\exp\left(\sum\int_{t_I^{u}}^{t}\lambda_{c',d'}^u(s)ds\right)}\\
=& p(c,d| t,\mathcal{H}_t^u)\times p(t|\mathcal{H}_t^{u})\\
=& p(c| t,\mathcal{H}_t^u)\times p(d|t,\mathcal{H}_t^u)\times p(t|\mathcal{H}_t^{u})\\
=& \frac{\lambda_{c}^u(t)}{\sum_{c'}\lambda_{c'}^u(t)} \times\frac{\lambda_{d}^u(t)}{\sum_{d'}\lambda_{d'}^u(t)}
\times p(t|\mathcal{H}_t^{u}),
\end{aligned}
\end{eqnarray}
where $p(t|\mathcal{H}_t^{u})$ is the conditional probability that there is an event happening at time $t$ given historical record, and $\lambda_{c,d}^u(t)$ measures the instantaneous happening rate of the event that the patient $u$ stay in a CU $d$ days before being transferred to the $c$-th CU. 
Focusing on the first two terms in the last row, we can find that the formulas of $p(c|t,\mathcal{H}_t^u)$ and $p(c|t,\mathcal{H}_t^u)$ in Eq.~(\ref{prob1}) are actually the normalized intensity functions.

Based on Eq.~(\ref{prob1}), we propose the following cross-entropy-based loss function for our learning task.
\begin{eqnarray}\label{cost}
\begin{aligned}
L(\bm{\Theta})=&-\sum_{u=1}^{U}\sum_{i=1}^{N^u}\biggl\{ 
\sum_{c=1}^{C}1\{c_i^u=c\}\log p(c| t_{i-1}^u, \mathcal{H}_{t_{i-1}^u}^u)\\
&+\sum_{d=1}^{D}1\{d_i^u=d\}\log p(d| t_{i-1}^u, \mathcal{H}_{t_{i-1}^u}^u)\biggr\}\\
=&-\sum_{u=1}^{U}\sum_{i=1}^{N^u}\log\left( \frac{\lambda_{c_i^u}^u(t_{i-1}^u)\lambda_{c_i^u}^u(t_{i-1}^u)}{\sum_{c'}\lambda_{c'}^u(t_{i-1}^u)\sum_{d'}\lambda_{d'}^u(t_{i-1}^u)}\right).
\end{aligned}
\end{eqnarray}
Here $1\{\mbox{statement}\}$ is an indicator of returning to 1 if the statement is truth, otherwise to 0.

Additionally, for exploring the relationship between the EHR-based feature and the patient flow, we consider the group sparsity of the parameter matrix of proposed model, denoted as $\|\bm{\Theta}\|_{1,2}$. 
$\|\bm{\Theta}\|_{1,2}=\sum_{m=1}^{M}\|\bm{\Theta}_m\|_2$ sums the $l_2$-norms of $\bm{\Theta}$'s rows $\bm{\Theta}_m$, $m=1,...,M$. 
\textcolor{black}{Here each dimension of feature is treated as a group.
Introducing this term as a regularizer into the loss function, we achieve feature selection simultaneously when learning model --- the rows corresponding to insignificant and noisy features will be suppressed to all zeros. 
Because the parameters of the model for predicting destination CUs and those for predicting durations are concatenated in $\bm{\Theta}$, the regularizer ensures that the useful features are shared via the two models.}
Such a feature selection strategy is also be used in~\cite{liu2009multi,yang20112}. 
In summary, we learn our discriminative point process model via solving the following optimization problem: 
\begin{eqnarray}\label{final}
\begin{aligned}
\min_{\bm{\Theta}}~L(\bm{\Theta}) + \gamma\|\bm{\Theta}\|_{1,2},
\end{aligned}
\end{eqnarray}
where $\gamma\geq 0$ is the weight controlling the significance of regularizer. 
Recalling the formula of conditional intensity function in Eq.~(\ref{expintensity}), we can easily find that Eq.~(\ref{final}) corresponds to a problem like multinomial logistic regression with group-lasso regularization~\cite{simon2013sparse}. 
From the viewpoint of Bayesian inference, the loss function $L(\bm{\Theta})$ corresponds to the negative log-likelihood function of $\bm{\Theta}$ given a series of samples, and the group-lasso regularizer imposes a structural prior distribution on $\bm{\Theta}$~\cite{raman2009bayesian,xu2015bayesian} such that the prior probability $p(\bm{\Theta})\propto \exp(-\gamma \sum_{m=1}^{M}\|\bm{\Theta}_m\|_2)$.

We apply the idea of alternating direction method of multipliers (ADMM)~\cite{gabay1976dual} to convert the optimization problem to several sub-problems that are easier to solve. 
Specifically, by introducing an auxiliary variable $\bm{X}$ and a dual variable $\bm{Y}$, we obtain the augmented Lagrangian of Eq.~(\ref{final}) as follows:
\begin{eqnarray*}\label{admm}
%\begin{aligned}
\min_{\bm{\Theta}}~L(\bm{\Theta}) + \gamma\|\bm{X}\|_{1,2}
+\rho \mbox{tr}(\bm{Y}^{\top}(\bm{\Theta}-\bm{X}))+\frac{\rho}{2}\|\bm{\Theta}-\bm{X}\|_F^2,
%\end{aligned}
\end{eqnarray*}
where $\rho>0$ is the penalty parameter. 
It mainly controls the convergence of ADMM algorithm~\cite{nishihara2015general}.  
$\mbox{tr}(\cdot)$ computes the trace of matrix. 
We solve it via optimizing the following sub-problems iteratively:

\textbf{Update $\bm{\Theta}$:} In the $k$-th iteration, we optimize the following problem: 
\begin{eqnarray*}
\begin{aligned}
\bm{\Theta}^{(k+1)}=\arg\min_{\Theta}L(\bm{\Theta})+\frac{\rho}{2}\|\bm{\Theta}-\bm{X}^{(k)}+\bm{Y}^{(k)}\|_F^2.
\end{aligned}
\end{eqnarray*}
Applying gradient descent algorithm, we update $\bm{\Theta}$ as
\begin{eqnarray}\label{updateTheta}
%\begin{aligned}
\bm{\Theta}^{(k+1)}=\bm{\Theta}^{(k)}-\beta
\nabla L|_{\bm{\Theta}^{(k)}}
-\beta\rho(\bm{\Theta}^{(k)}-\bm{X}^{(k)}+\bm{Y}^{(k)}),
%\end{aligned}
\end{eqnarray}
where parameter $\beta>0$ is the learning rate for updating parameters. 
$\nabla L|_{\bm{\Theta}^{(k)}}$ is the gradient of loss function $L(\bm{\Theta}^{(k)})$ given current parameters $\bm{\Theta}^{(k)}$, which is computed as
\begin{eqnarray*}
\begin{aligned}
&\nabla L|_{\bm{\theta}_c^{(k)}}=\sum_{u=1}^{U}\sum_{i=1}^{N^u}\left( \frac{\lambda_c^{u,(k)}(t_{i-1}^u)}{\sum_{c'}\lambda_{c'}^{u,(k)}(t_{i-1}^u)}-1\{c_i^u=c\} \right)\bm{f}_{t_{i-1}}^u,\\
&\nabla L|_{\bm{\theta}_d^{(k)}}=\sum_{u=1}^{U}\sum_{i=1}^{N^u}\left( \frac{\lambda_d^{u,(k)}(t_{i-1}^u)}{\sum_{d'}\lambda_{d'}^{u,(k)}(t_{i-1}^u)}-1\{d_i^u=d\}\right)\bm{f}_{t_{i-1}}^u.
\end{aligned}
\end{eqnarray*}
Here $\lambda_c^{u,(k)}(t)$ and $\lambda_d^{u,(k)}(t)$ are estimates of conditional intensity functions given current parameters. 

\textbf{Update $\bm{X}$:} The optimization problem is a simple linear model with group-lasso penalty~\cite{simon2013sparse,mosci2010solving,parikh2014proximal}:
\begin{eqnarray*}
\begin{aligned}
\bm{X}^{(k+1)}=\arg\min_{\bm{X}}&\frac{\rho}{2}\|\bm{\Theta}^{(k+1)}-\bm{X}+\bm{Y}^{(k)}\|_F^2+\gamma\|\bm{X}\|_{1,2}.
\end{aligned}
\end{eqnarray*}
Denote $\bm{X}_m$ as the $m$-th row of $\bm{X}$. Its subgradient equations are 
\begin{eqnarray}\label{subgrad}
\begin{aligned}
\rho(\bm{X}_m-(\bm{\Theta}_m^{(k+1)}+\bm{Y}_m^{(k)}))+\gamma\bm{s}=0,
\end{aligned}
\end{eqnarray} 
where $\bm{s}=\frac{\bm{X}_m}{\|\bm{X}_m^{(k)}\|_2}$ if $\bm{X}_m^{(k)}\neq \bm{0}$ and $\bm{s}$ is a vector with $\|\bm{s}\|_2<1$ otherwise. 
The solution of Eq.~(\ref{subgrad}) is 
\begin{eqnarray*}
\begin{aligned}
\hat{\bm{X}}_{m}=\left(1+\frac{\rho}{\gamma \|\bm{X}_m^{(k)}\|_2}\right)^{-1}(\bm{\Theta}_m^{(k+1)}+\bm{Y}_m^{(k)}),
\end{aligned}
\end{eqnarray*}
and then, $\bm{X}_m^{(k+1)}$ is updated via
\begin{eqnarray}\label{updateX}
\begin{aligned}
\bm{X}_m^{(k+1)}=
\begin{cases}
\bm{0}, &\mbox{if}~\|\hat{\bm{X}}_m-(\bm{\Theta}_m^{(k+1)}+\bm{Y}_m^{(k)})\|_2\leq \frac{\gamma}{\rho},\\
\hat{\bm{X}}_m, &\mbox{otherwise}.
\end{cases}
\end{aligned}
\end{eqnarray}
\begin{eqnarray}\label{updateY}
\begin{aligned}
\mbox{\textbf{Update $\bm{Y}$:}}~\bm{Y}^{(k+1)}=\bm{Y}^{(k)}+(\bm{\Theta}^{(k+1)}-\bm{X}^{(k+1)}).
\end{aligned}
\end{eqnarray}
Repeating the steps above until convergence, we learn the parameter matrix of the model, and obtain $p(c|t, \mathcal{H}_t^u)$ and $p(d|t,\mathcal{H}_t^u)$ jointly. 
In summary, we give the scheme of our learning algorithm in Algorithm~\ref{alg1}.

\begin{algorithm}[h!]
%\vspace{-2pt}
  \caption{Discriminative Learning of Mutually-Correcting Processes (DMCP)}
  \label{alg1}
  \begin{algorithmic}
    \Require Patient flow $\{s_u\}_{u=1}^{U}$, parameters $\gamma, \rho, \beta$, error bound $\epsilon=0.01$.
    \Ensure $\bm{\Theta}$.
    \State Initialize $\bm{\Theta}^{(0)}$ randomly, $\bm{X}^{(0)}=\bm{\Theta}^{(0)}$, $\bm{Y}^{(0)}=\bm{0}$, outer iteration number $k = 0$
    \Repeat
    		\State Inner iteration number $l=0$, $\bm{\Theta}^{(k,l)}=\bm{\Theta}^{(k)}$.
    		\Repeat
    			\State Update $\bm{\Theta}^{(k,l+1)}$ via Eq.~(\ref{updateTheta}).
    			\State $l=l+1$.
    		\Until{$\frac{\|\bm{\Theta}^{(k,l)}-\bm{\Theta}^{(k,l-1)}\|_2}{\|\bm{\Theta}^{(k,l)}\|_2}\leq\epsilon$}
    		\State $\bm{\Theta}^{(k+1)}=\bm{\Theta}^{(k,l)}$.
         \State Update $\bm{X}^{(k+1)}$ via Eq.~(\ref{updateX}).
         \State Update $\bm{Y}^{(k+1)}$ via Eq.~(\ref{updateY}).
         \State $k=k+1$.
    \Until{$\frac{\|\bm{\Theta}^{(k)}-\bm{\Theta}^{(k-1)}\|_2}{\|\bm{\Theta}^{(k)}\|_2}\leq\epsilon$.}
    \State $\bm{\Theta}=\bm{\Theta}^{(k)}$.
  \end{algorithmic}
\end{algorithm}

Our model and algorithm can be viewed as a trade-off between learning joint probability $p(c,d|t,\mathcal{H}_t^u)$ directly and learning the probabilities of transition and duration ($p(c|t,\mathcal{H}_t^u)$ and $p(d|t,\mathcal{H}_t^u)$) independently. 
On one hand, learning $p(c,d|t,\mathcal{H}_t^u)$ requires $\mathcal{O}(CD)$ parameters, which might lead to the over-fitting result.
Our model, however, merely requires $\mathcal{O}(C+D)$ parameters.

On the other hand, although we relax the weak correlation between the transition and the duration to an independence assumption, we do not really learn $p(c|t,\mathcal{H}_t^u)$ and $p(d|t,\mathcal{H}_t^u)$ independently. 
With the help of the group-lasso in Eq.~(\ref{final}), their correlation is preserved to some degree --- the group sparsity of parameters is shared via $p(c|t,\mathcal{H}_t^u)$ and $p(d|t,\mathcal{H}_t^u)$ and the parameters are updated simultaneously. 

It should be noted that our discriminative algorithm is not only suitable for mutually-correcting processes. 
Actually, we can use conditional intensity functions from arbitrary point processes to compute the conditional probabilities in Eq.~(\ref{prob1}) and the loss function in Eq.~(\ref{cost}).

\subsection{Enhancing Robustness to Imbalanced Data}\label{sec:ub}
\begin{figure}[!ht]
\centering
\subfigure[]{
\includegraphics[width=0.22\linewidth]{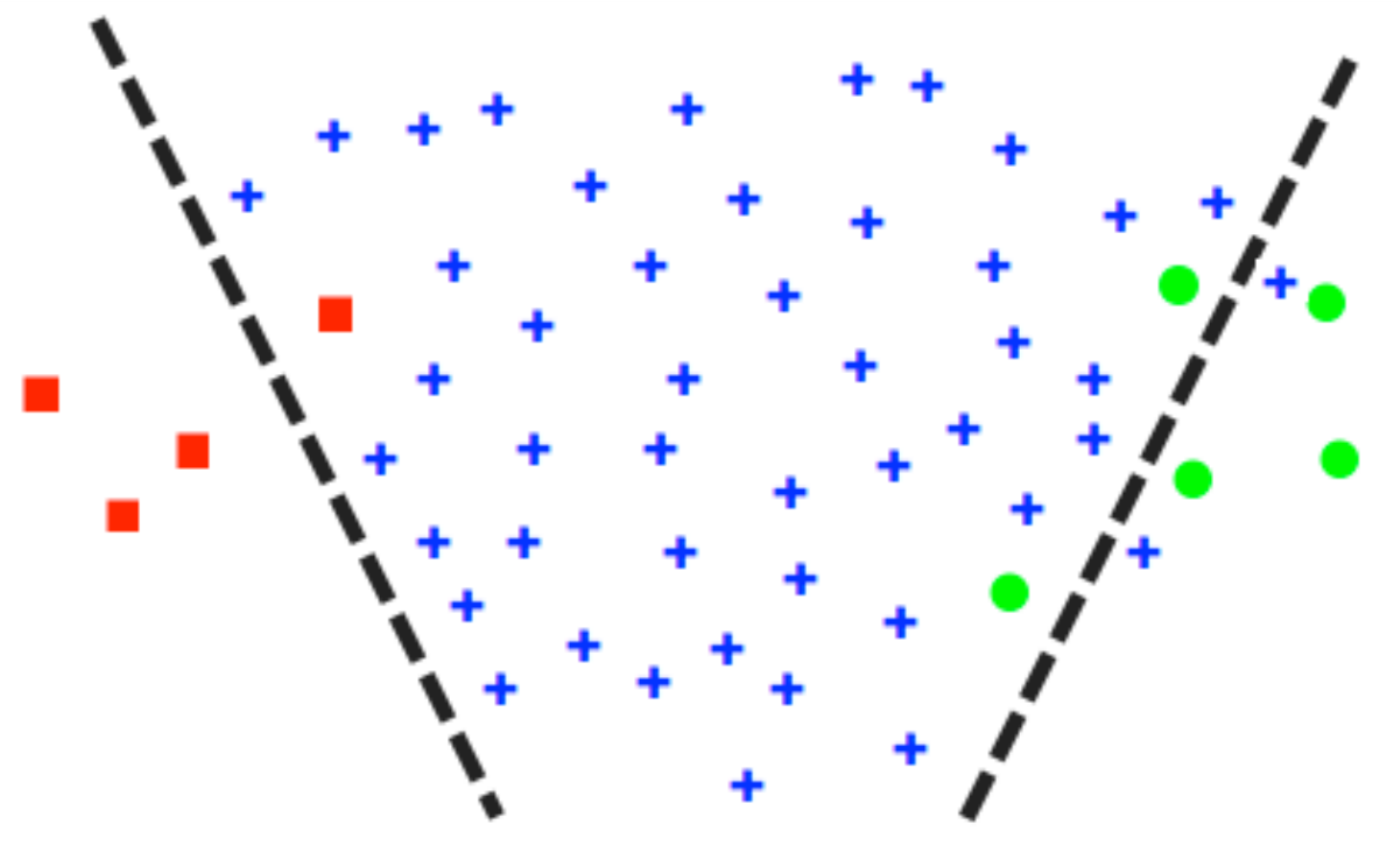}\label{FigUb1}
}~
\subfigure[]{
\includegraphics[width=0.22\linewidth]{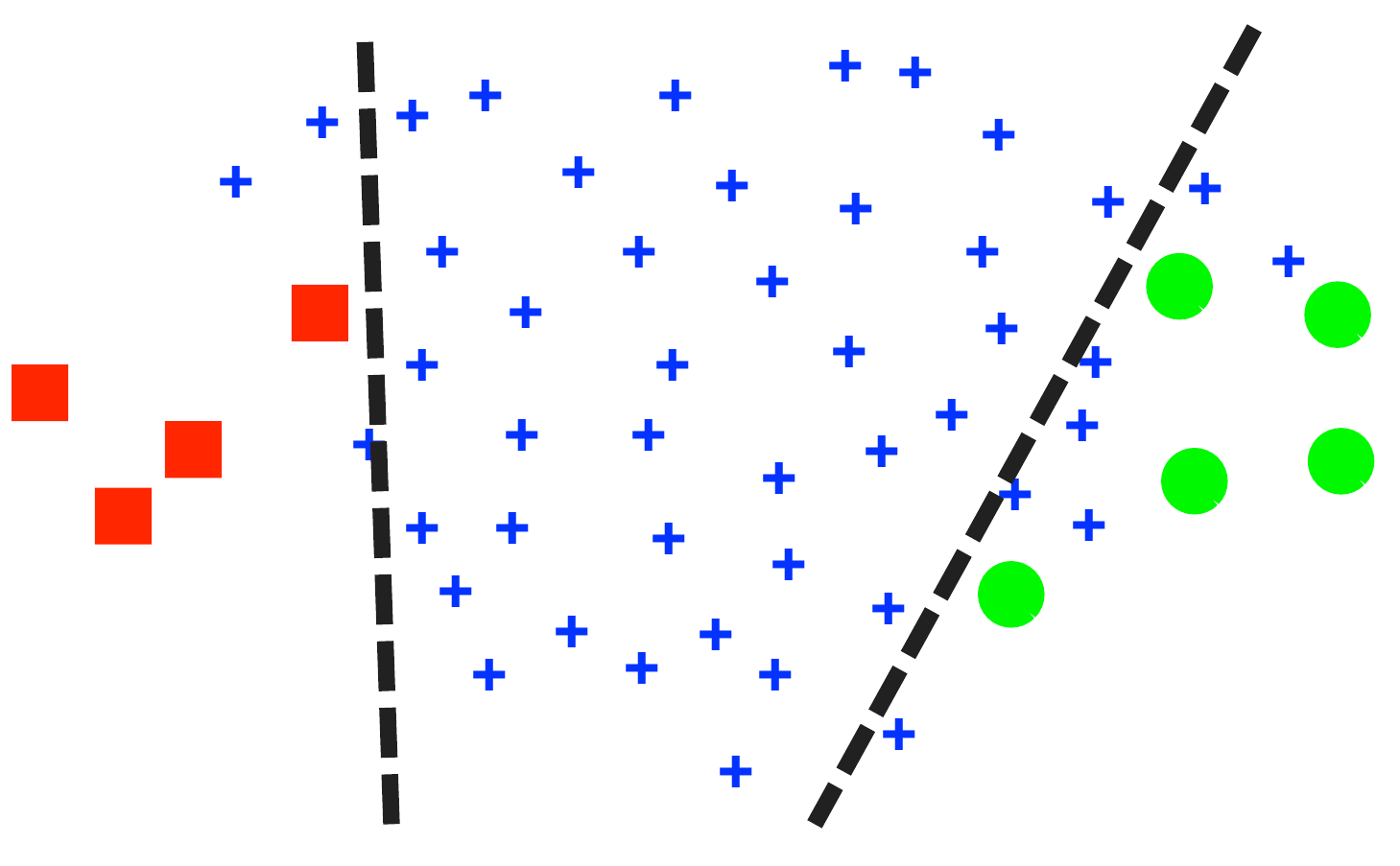}\label{FigUb2}
}~
\subfigure[]{
\includegraphics[width=0.22\linewidth]{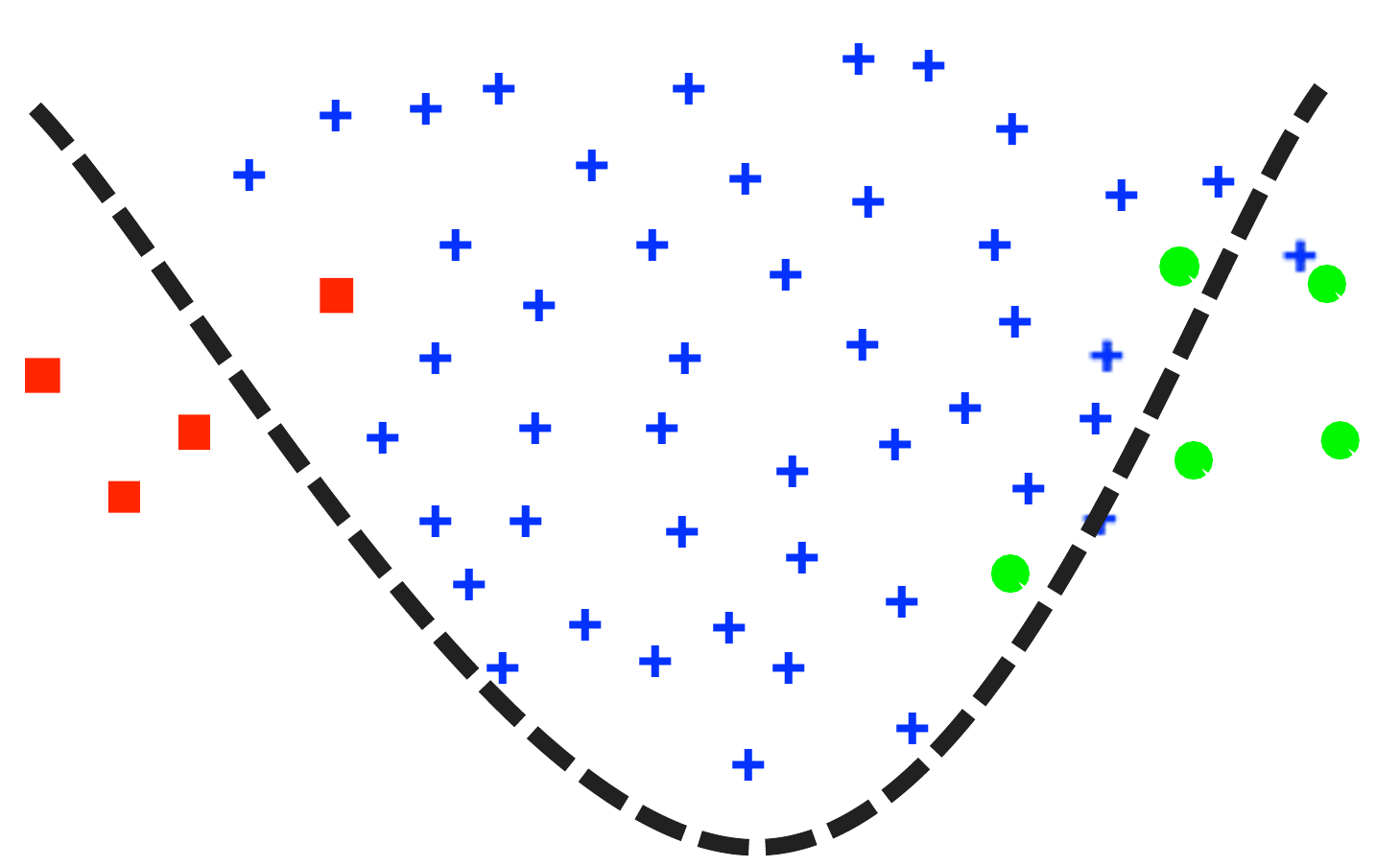}\label{FigUb3}
}~
\subfigure[]{
\includegraphics[width=0.22\linewidth]{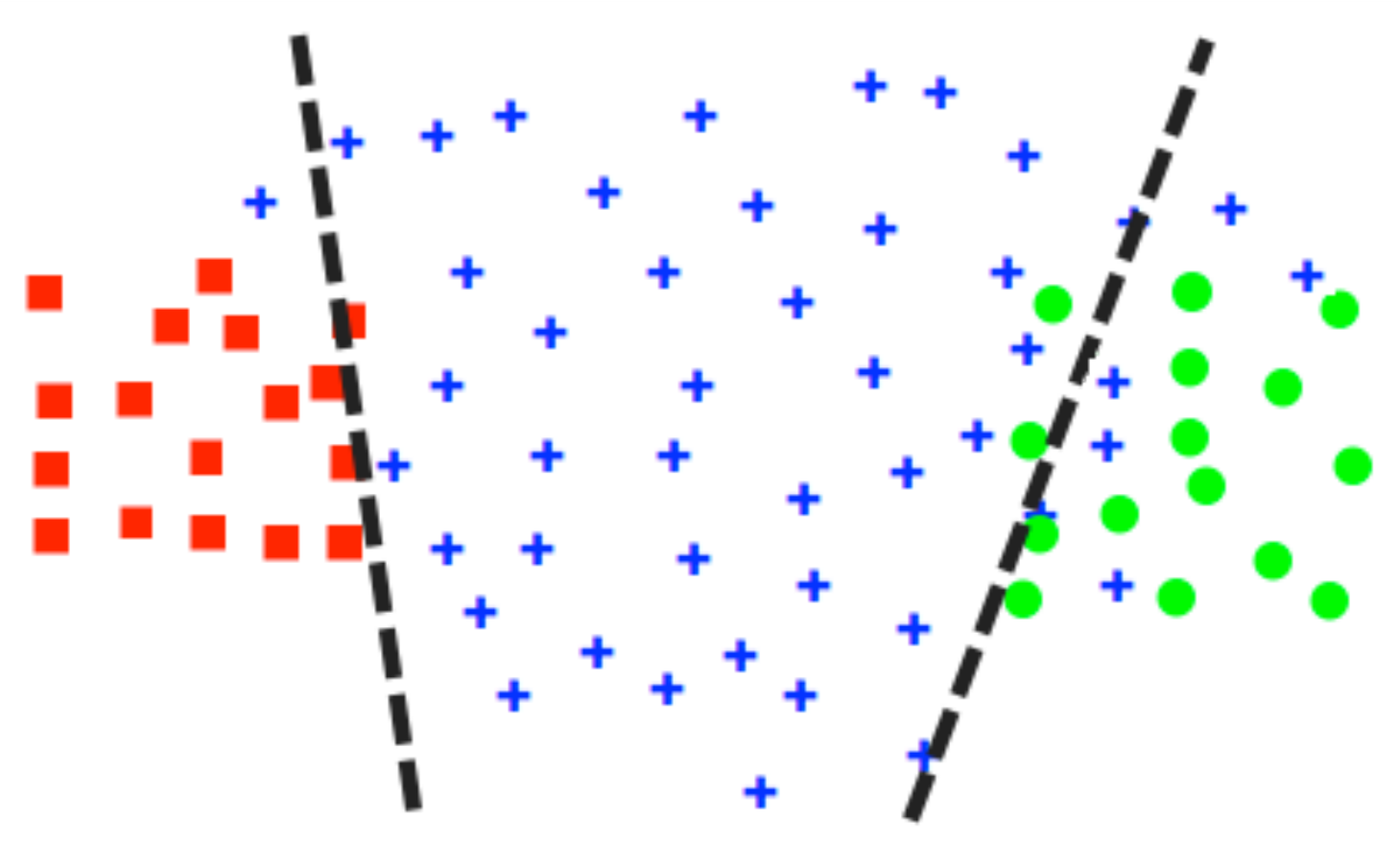}\label{FigUb4}
}
\caption{
The simple 2-D example illustrating various methods to solve data imbalance problem: (a) The original data having 3 classes is shown, where blue crosses are samples of major class while red squares and green dots are samples of two minor classes, respectively. 
(b) The weighed data is shown, where the samples of minor classes have large weights (enlarged). 
(c) The hierarchical data is shown, where the samples are unchanged while a nonlinear binary classifier is learned. 
(b) The synthetic data is shown, where the minor classes are supplemented via auxiliary samples. 
The classifiers in the subfigures are shown as black dotted lines and curves.}\label{Figimbalance}
\end{figure}

As aforementioned, the imbalance of the data has a remarkable impact on the overall performance of patient work flow prediction, leading to the poor performance of duration and transition prediction of classes with minority samples (i.e., in the following experiments, the prediction accuracy of destination CUs with only a few patients transferring to CUs like ACU, FICU, TSICU, is relatively lower than other CUs with more patients like CCU, SCRU, MICU, NICU).
As the 2-D case in Fig.~\ref{FigUb1} shows, the classifier trained on imbalanced data will focus more on the classification accuracy of the class having sufficient samples while ignore the errors of the class having extremely few samples. 

For suppressing the negative influence of data imbalance problem, several potential solutions are proposed and analyzed in depth. 

\textbf{Weighted data.} A reason of the low prediction accuracy of the classes with few samples is that these classes are insignificant compared to the the classes with sufficient samples when we optimize the likelihood or loss  function of the classifier. 
A possible way to increase the significance of the classes with few samples is adding the weights of the samples in the training phase~\cite{king2001logistic,tan2005neighbor,lee2003learning,zhuang2005efficient}. 
Specifically, we can rewrite the likelihood function in Eq.~(\ref{cost}) as
%\begin{eqnarray*}
%\begin{aligned}
$-\sum_{u=1}^{U}\sum_{i=1}^{N^u}w_i\log\left( \frac{\lambda_{c_i^u}^u(t_{i-1}^u)\lambda_{c_i^u}^u(t_{i-1}^u)}{\sum_{c'}\lambda_{c'}^u(t_{i-1}^u)\sum_{d'}\lambda_{d'}^u(t_{i-1}^u)}\right)$, 
%\end{aligned}
%\end{eqnarray*}
where the weight $w_i$ aims at suppressing the imbalance of data. 
It should be large for the samples in the minor classes and small for those in the major ones (i.e., in our case, counting the number of labels $\{(c, d)\}$ in the training set, denoted as $\#\{(c,d)\}$, we calculate $w_i=\frac{1}{\log(1+\#\{(c,d)\})}$ if $c_i^u=c$ and $d_i^u=d$). 
Fig.~\ref{FigUb2} visualizes the weighted data, where the enlarged squares and dots are the samples with large weights. 
Such a simple method might increase the classification accuracy of the classes with few training samples, while decrease the classification accuracy of the classes with sufficient samples at the same time. 
Because simply weighting samples may wrongly change the distribution of samples in minor classes, the outliers of the classes are enhanced and wrong boundaries between classes are learned.

\textbf{Hierarchical data.}
Instead of learning one multi-class classifier directly with imbalanced data, we can rank classes according to the number of training samples and learn binary classifiers hierarchically~\cite{sethi1982hierarchical,longford1987fast}. 
Specifically, in each step, we take the class with the largest number of training samples as ``MAJORITY'', and the rest samples as a single class called ``MINORITY''. 
Then, a binary classifier is trained on them and the samples of ``MAJORITY'' is removed from the training set. 
Repeating the steps above, we obtain a series of binary classifier from hierarchical data. 
The principle of this method is re-balancing data via merging minor classes. 
However, in practice, the merging step may lead the classes to be linear-inseparable, which increase the difficulty of training phase.
In this case, as Fig.~\ref{FigUb3} shows, nonlinear binary classifier is required in each step, which relies on more complicated learning algorithm, e.g., kernel-based methods. When training linear classifier insistently, the classification accuracy may not be improved. 	

\textbf{Synthetic data.}
For overcoming the weaknesses of the two methods above, we propose a new method to solve the data imbalance problem. 
Recalling the classifier trained from weighted data, we can view the weighted data as sampling minor class repeatedly and generating identical samples. 
Different from directly sampling identical samples, we propose a data synthesis method: for the samples (feature vectors) in a minor class, we synthesize auxiliary samples for the class by sampling each element according to the distribution of corresponding elements of existing samples. 
Therefore, the auxiliary samples are similar but not identical to original ones. 
Supplementing these auxiliary samples to the minor classes as training samples\footnote{The numbers of samples in different classes are equal after the pre-processing.}, as shown in Fig.~\ref{FigUb4}, we can enhance the robustness of the learning method to imbalanced data. 

\textcolor{black}{Our data synthesis method is actually based on an assumption that the dimensions of feature are independent with each other. 
As long as the assumption is held by original data, our method can guarantee that the auxiliary samples yields to the distribution of original data. 
On the contrary, the two competitors mentioned above change the distribution of data: the weighted data implicitly increases the probability of those samples in minor classes; the hierarchical data also changes the distribution of minor classes in each step. 
As a result, the models learned based on the data generated via these two methods have higher risk of model misspecification.} 
In the following experiments, we will show that applying our data synthesis method as a pre-processing in the training phase, we can enhance the robustness of learning method and obtain superior testing results to its competitors.

\subsection{Patient Flow Prediction} 
Given learned model $\bm{\Theta}$, we can predict patient flow for each patient $u$ simply. 
Specifically, given historical record $\mathcal{H}_{t_{i-1}}$, we compute $p(c|t_{i-1}^u, \mathcal{H}_{t_{i-1}}^u)$ and $p(d|t_{i-1}^u, \mathcal{H}_{t_{i-1}}^u)$ for $c\in\mathcal{C}$ and $d\in\mathcal{D}$, respectively. 
The predicts of $c_i^u$ and $d_i^u$ are given as
\begin{eqnarray*}
\begin{aligned}
&\hat{c}_i^u=\arg\max_{c\in\mathcal{C}}p(c|t_{i-1}^u, \mathcal{H}_{t_{i-1}}^u),\\
&\hat{d}_i^u=\arg\max_{d\in\mathcal{D}}p(d|t_{i-1}^u, \mathcal{H}_{t_{i-1}}^u).
\end{aligned}
\end{eqnarray*}

\section{Experiments}\label{sec:exp}

\subsection{Baselines and Evaluations}
Although there is no existing method proposed to predict patient flow based on a large amount of EHRs, we consider several alternatives that can be potentially adapted to solve our problem. 
These potential methods are designed for modeling transition processes in  discrete or continuous time. 
Taking these methods as baselines, we compare our method (DMCP) with them and demonstrate its superiority.

\textbf{Markov chain (MC).} Taking $C$ CUs and $D$ duration days as states, the simplest method is treating the event sequences as two independent Markov chains for the transition and the duration, respectively. 
Two one-order MCs are trained, whose transition matrices are calculated via counting the transitions among various states. In the prediction phase, given initial state (i.e., current CU and previous duration time), we use the transition matrices to predict next states (i.e., current duration time and next CU). 

\textbf{Vector auto-regressive model (VAR).} Similar to the MC model, the VAR model used in this paper also captures the transitions among CUs and the durations in CUs as two independent transition processes, whose transition matrices are learned via the method in~\cite{han2013transition}. 
Different from the MC model, the transition matrix of the VAR model does not have probabilistic interpretation but is more flexible.

\textbf{Continuous-time Markov Chain (CTMC).} The CTMC~\cite{anderson2012continuous}, as a special type of semi-Markov model~\cite{krol2015semimarkov}, also models the transition among CUs as a markov process. 
In this application, the transition process among CUs is modeled as a Markov chain in continuous time, whose transition probability is time-varying. 
In the prediction phase, the destination CU is predicted according to previous CU and current transition matrix, and the duration in current CU is predicted via the interval between adjacent transitions. 

\textbf{Logistic regression (LR).} Using the feature extracted from EHRs, we can treat the prediction of CU patient flow as a classification problem. 
Specifically, two multi-class classifiers are trained independently via multinomial logistic regression (or called softmax regression) for destination CUs and duration days, respectively. 
In the training set, for each label $c_i^u$ (or $d_i^u$), the feature is $[\bm{f}_0^{u\top}, \bm{f}_i^{u\top}]^{\top}$. 

\textbf{Hawkes processes (HP).} Taking the transitions among CUs as event sequences, the parametric Hawkes process model~\cite{li2014learning} is implemented, where the conditional intensity function is shown in Table~\ref{tab:PP}. 
Different from our method, the Hawkes process is learned in a generative way --- the likelihood of the whole event sequence is maximized via the maximum likelihood estimator (MLE), i.e., $\max_{\bm{\Theta}}\prod_u\prod_i p(c_i^u, d_i^u |\mathcal{H}_t^u)(1-P(T^u))$. 
In the prediction phase, given historical record $\mathcal{H}_t^u$, we compute the intensity of CUs, $\lambda_c^u(t)$, is computed in the time interval $[t, t+D]$, the predictions of next event $(c, d)$ are obtained via $\max_{(c,d)\in \mathcal{C}\times\mathcal{D}}\int_{t+d-1}^{t+d}\lambda_c^u(s)ds$.

\textbf{Modulated Poisson processes (MPP).} The MPP method replaces our mutually-correcting process with the modulated Poisson process shown in Table~\ref{tab:PP}. In the learning phase, the multinomial logistic regression is applied as we did while the group-lasso is not considered. 
From the viewpoint of methodology, this method can be viewed as a point process-based interpretation of the generalized logit model of Markov chain in~\cite{cole2005multistate}.

\textbf{Self-correcting process (SCP).} Similar to the MPP method, the SCP method replaces our mutually-correcting process with the self-correcting process shown in Table~\ref{tab:PP}. In the learning phase, the multinomial logistic regression is applied as we did while the group-lasso is not considered.

The baselines above can be categorized into three classes: the MC, VAR, and CTMC methods are feature-independent, which merely rely on temporal information; the LR is history-independent, which merely relies on the EHR-based feature generated at current time while ignores historical record; the HP, MPP, SCP, including our DMCP are point process-based methods. 
Specifically, the MPP, SCP, and our DMCP can be viewed as extensions of the LR method, which merge current features with historical ones via various point process models. 
Additionally, our method is the only one introducing group-lasso into learning algorithm. 

For evaluating the significance and the performance of pre-processing of imbalanced data, we consider our DMCP method with various pre-processing methods, including the \textbf{weighted data+DMCP (WDMCP)}, the \textbf{hierarchical data+DMCP (HDMCP)}, and the proposed \textbf{synthetic data+DMCP (SDMCP)}. 
The SCP with synthetic data \textbf{SSCP} is also tested to prove the universality of our pre-processing method.

%\textcolor{black}{
Using the proposed data representation method, we can extract a large amount of feature-label pairs from event sequences, e.g., $(f_{t_{i-1}}^u, c_i^u, d_i^u)$, where $f_{t_{i-1}}^u$ is the feature of patient $u$ containing her historical information before time $t_{i-1}$, $c_i^u$ is her destination CU after $t_{i-1}$, and $d_i^u$ is the duration time in $c_i^u$ accordingly. 
Given all these pairs, we train and test all the methods via 10-fold cross validation. 
Specifically, we use 90\% of the data for training and the remaining 10\% for testing randomly. 
The training data is further divided into 10 folds. 
For each method, its model is trained via 10 trials. 
In each trial, the 9-fold data is used to train the model while the rest is for validation.
The final model is the average of 10 training results.%}
 
For evaluating various methods comprehensively, we apply the following measurements:

\textbf{Prediction accuracy:} The prediction accuracy $\mbox{AC}_{c}$ for each CU $c$ and the overall accuracy $\mbox{AC}_C$ are calculated as
\begin{eqnarray*}
\begin{aligned}
\mbox{AC}_c=\frac{\mbox{\#$\{$right prediction$\}$}}{\mbox{ \#$\{$transitions to $c\}$}},~
\mbox{AC}_C=\sum_{c=1}^C\frac{\mbox{\#$\{$transitions to $c\}$}}{\mbox{ \#$\{$total transitions$\}$}}\mbox{AC}_c. 
\end{aligned}
\end{eqnarray*}
The prediction accuracy $\mbox{AC}_{d}$ for each duration category $d$ and the overall accuracy $\mbox{AC}_D$ are calculated in the same way.

\textbf{Relative simulation error:} Given trained model, we can simulate patient flow following existing data. 
Specifically, given historical patient data, we simulate the daily number of patients in each CU within the following week. 
The relative simulation error of patient flow $\mbox{Err}_c$ for each CU $c$ and the overall relative error $\mbox{Err}_C$ are calculated as
\begin{eqnarray*}
\begin{aligned}
\mbox{Err}_c=\frac{1}{7}\sum_{d=1}^{7}\frac{|N_{c,d}-\hat{N}_{c,d}|}{N_{c,d}},~
\mbox{Err}_C=\frac{1}{7}\sum_{d=1}^{7}\frac{|N_{d}-\hat{N}_{d}|}{N_{d}},
\end{aligned}
\end{eqnarray*}
where $N_{c,d}$ ($N_d$) is the real number of patient in each CU (all CUs) in the $d$-th day, and $\hat{N}_{c,d}$ ($\hat{N}_d$) is the simulation result. 

It should be noted that we also try to learn joint probability $p(c,d|t,\mathcal{H}_t^u)$ directly. 
As we analyzed in the end of section 3.3, such a method will lead to serious over-fitting problem --- even on the pre-processed data, the prediction accuracy for each $(c,d)$ pair is no more than $0.31$, and the simulation error is larger than $0.45$. 
Compared with the result of our method shown below, the performance is too bad to be applicable. 

\textbf{The robustness of algorithm to parameters:} The influences of parameters on our learning algorithm are investigated. 
Specifically, we give a strategy for selecting learning rate $\beta$ and analyze the function of the bandwidth of Gaussian kernel $\sigma$ in our mutually-correcting process model. 
The weight of regularizer $\gamma$ and the weight of augmented Lagrangian $\rho$ are also analyzed.

\begin{figure}[!t]
\centering
\subfigure[The prediction accuracy of destination CUs obtained via various methods.]{
\includegraphics[width=0.98\linewidth]{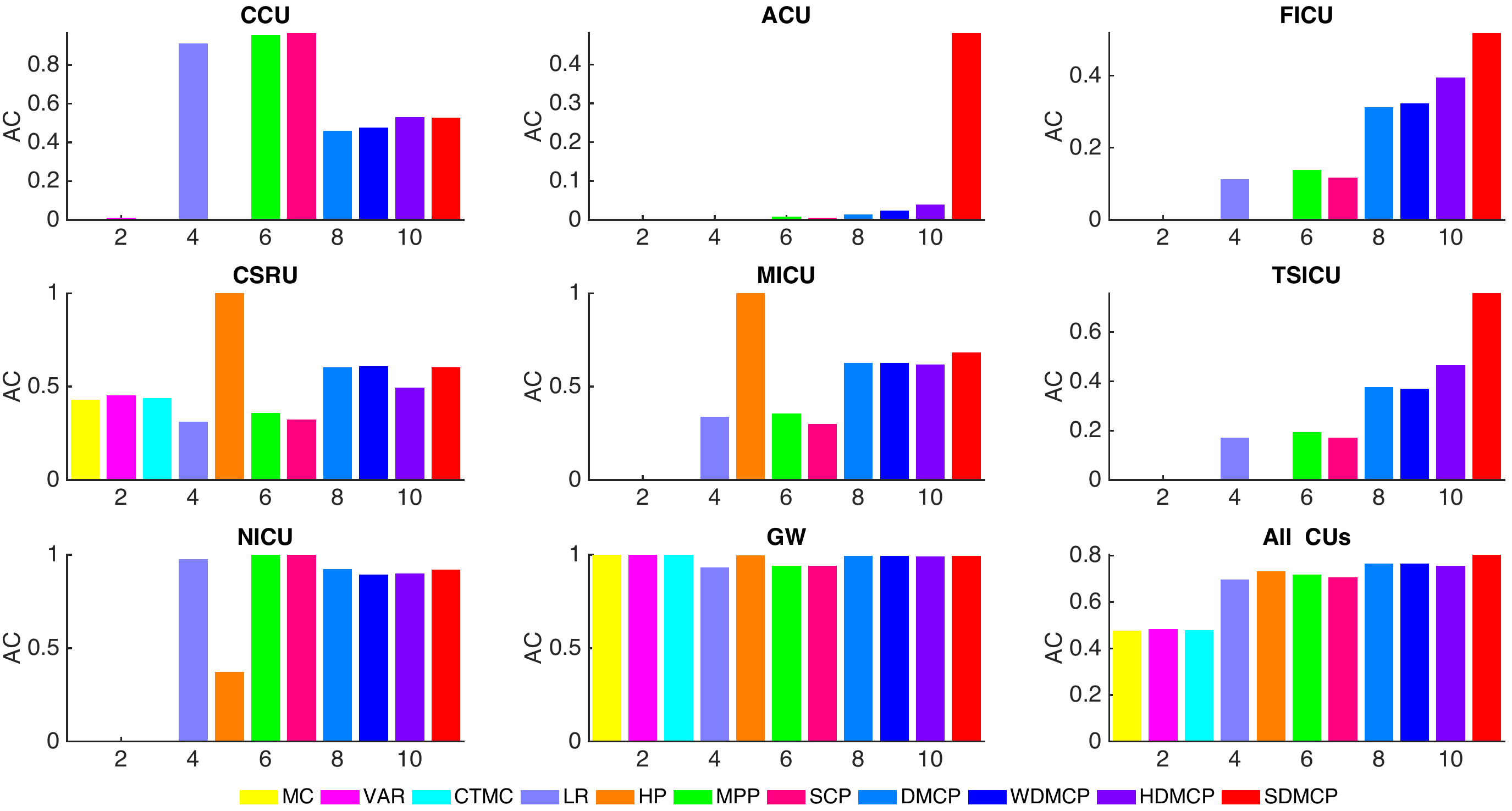}\label{FigAccC}
}\\
\subfigure[The prediction accuracy of duration days obtained via various methods.]{
\includegraphics[width=0.98\linewidth]{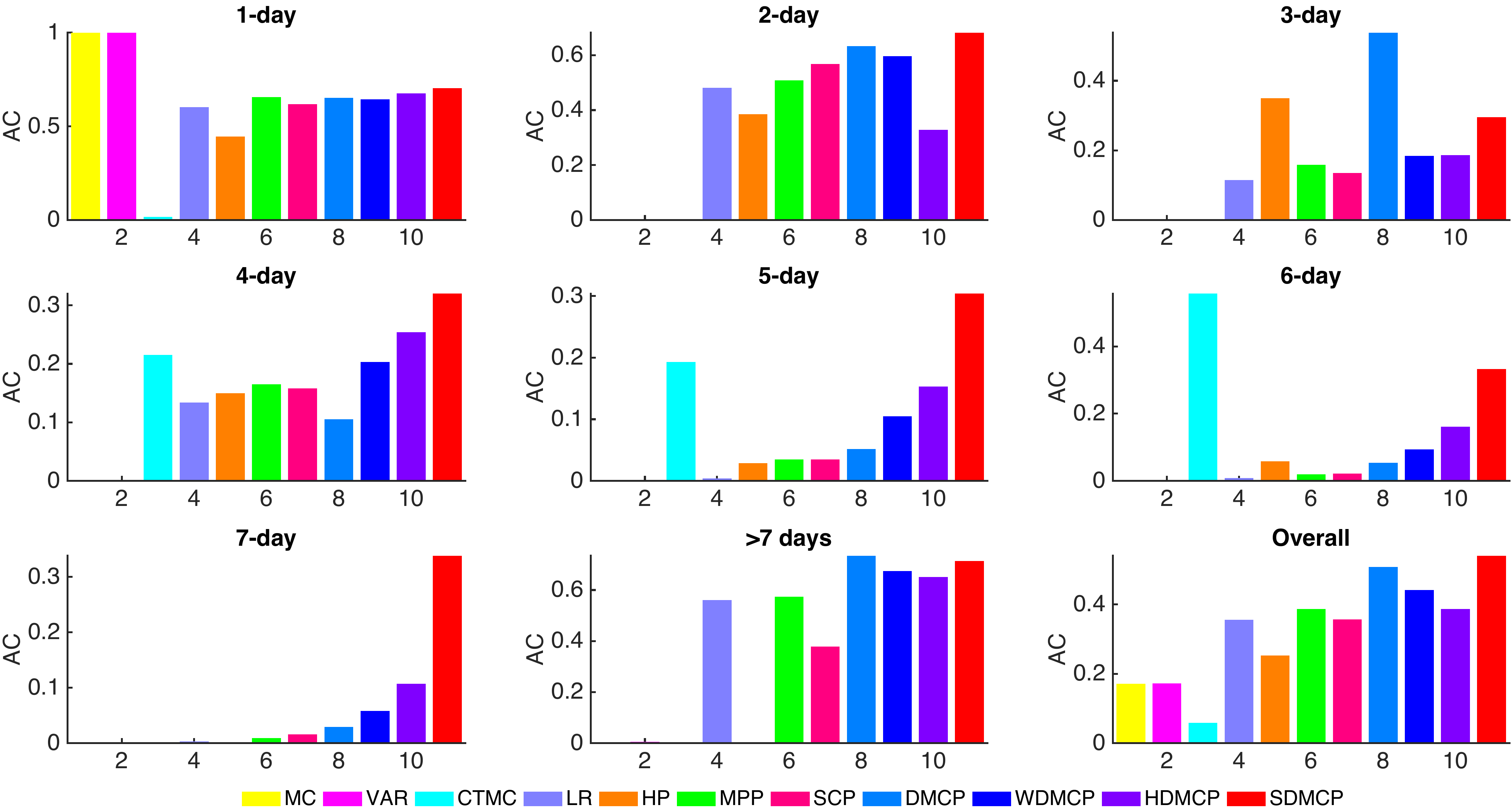}\label{FigAccD}
}
\caption{
(a) The prediction accuracy of each CUs and the overall accuracy are given. 
In each subfigure, the color bars correspond to various learning methods. 
(b) The prediction accuracy of each duration day and the overall accuracy are given. 
In each subfigure, the color bars correspond to various learning methods.}\label{FigAcc}
\end{figure}

\begin{figure}[!t]
\centering
\includegraphics[width=0.98\linewidth]{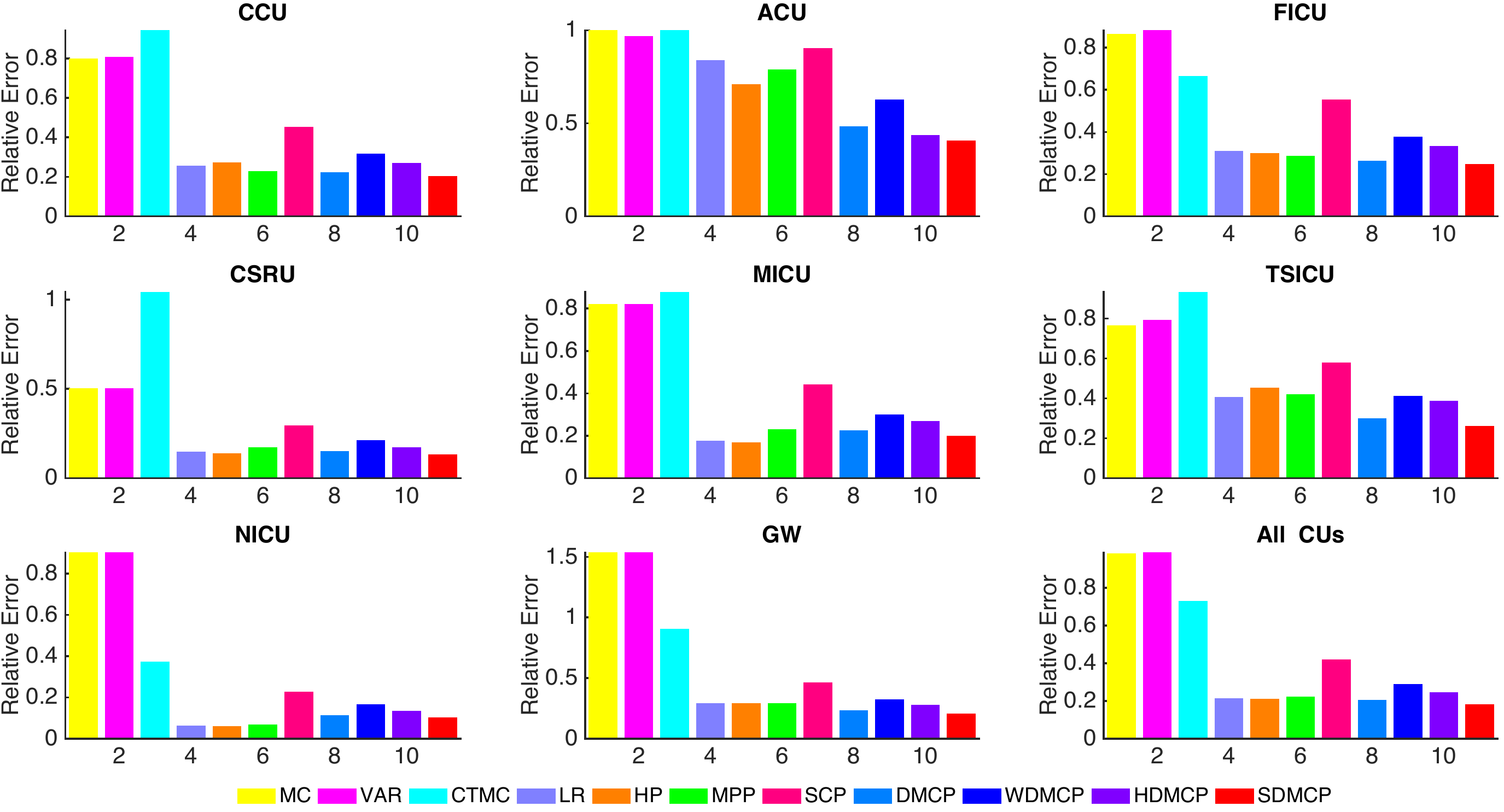}
\caption{
The relative simulation error of each CU and the overall simulation error are given. 
In each subfigure, the color bars correspond to various learning methods.}\label{FigSim}
\end{figure}

\begin{table*}[t!]
  \centering
  \small\caption{Prediction accuracy for various methods on destination CUs}\label{tab:acc}
  \begin{threeparttable}[c]
      \begin{tabular}{
        @{\hspace{6pt}}c|c|c@{\hspace{6pt}}
        @{\hspace{6pt}}c@{\hspace{6pt}}
        @{\hspace{6pt}}c@{\hspace{6pt}}
        @{\hspace{6pt}}c@{\hspace{6pt}}
        @{\hspace{6pt}}c@{\hspace{6pt}}
        @{\hspace{6pt}}c@{\hspace{6pt}}
        @{\hspace{6pt}}c@{\hspace{6pt}}
        @{\hspace{6pt}}c@{\hspace{6pt}}
        @{\hspace{6pt}}|c@{\hspace{6pt}}
        @{\hspace{6pt}}c@{\hspace{6pt}}
        @{\hspace{6pt}}c@{\hspace{6pt}}
        @{\hspace{6pt}}c@{\hspace{6pt}}
        }
        \hline\hline
       \multicolumn{2}{c|}{\multirow{2}{*}{Method}}&\multicolumn{8}{c}{Without pre-processing} & \multicolumn{4}{|c}{With pre-processing}\\ \cline{3-14}
\multicolumn{2}{c|}{}&MC  &VAR  &CTMC  &LR   &HP   &MPP   &SCP   &DMCP &SSCP &WDMCP &HDMCP &SDMCP\\ \hline
\multirow{8}{*}{$\mbox{AC}_c$'s}&CCU    &0   &0.010&0     &0.912&0    &0.954 &0.965 &0.461&\textbf{0.979}&0.479 &0.534 &0.529\\
&ACU    &0   &0    &0     &0.002&0    &0.008 &0.006 &0.014&0.295&0.025 &0.040 &\textbf{0.482}\\
&FICU   &0   &0.002&0     &0.113&0    &0.138 &0.117 &0.313&0.393&0.325 &0.396 &\textbf{0.520}\\
&CSRU   &0.430&0.455&0.438&0.311&\textbf{0.999}&0.359 &0.323 &0.605&0.486&0.612 &0.496 &0.606\\
&MICU   &0   &0    &0     &0.338&\textbf{0.997}&0.356 &0.301 &0.628&0.470&0.629 &0.619 &0.684\\
&TSICU  &0   &0    &0     &0.172&0    &0.194 &0.171 &0.376&0.525&0.371 &0.466 &\textbf{0.758}\\
&NICU   &0   &0.005&0     &0.977&0.372&0.995 &\textbf{0.997} &0.924&0.486&0.894 &0.901 &0.920\\
&GW     &1.000&1.000&1.000&0.932&\textbf{0.996}&0.942 &0.940 &0.995&0.963&0.995 &0.991 &0.995\\ \hline
$\mbox{AC}_C$&All CUs&0.478&0.483&0.479&0.696&0.731&0.719 &0.705 &0.766&0.724&0.766 &0.758 &\textbf{0.805}\\
        \hline\hline
      \end{tabular}
   \end{threeparttable}
\end{table*}

\begin{table*}[t!]
  \centering
  \small\caption{Prediction accuracy for various methods on duration days}\label{tab:acd}
  \begin{threeparttable}[c]
      \begin{tabular}{
        @{\hspace{6pt}}c|c|c@{\hspace{6pt}}
        @{\hspace{6pt}}c@{\hspace{6pt}}
        @{\hspace{6pt}}c@{\hspace{6pt}}
        @{\hspace{6pt}}c@{\hspace{6pt}}
        @{\hspace{6pt}}c@{\hspace{6pt}}
        @{\hspace{6pt}}c@{\hspace{6pt}}
        @{\hspace{6pt}}c@{\hspace{6pt}}
        @{\hspace{6pt}}c@{\hspace{6pt}}
        @{\hspace{6pt}}|c@{\hspace{6pt}}
        @{\hspace{6pt}}c@{\hspace{6pt}}
        @{\hspace{6pt}}c@{\hspace{6pt}}
        @{\hspace{6pt}}c@{\hspace{6pt}}
        }
        \hline\hline
       \multicolumn{2}{c|}{\multirow{2}{*}{Method}}&\multicolumn{8}{c}{Without pre-processing} & \multicolumn{4}{|c}{With pre-processing}\\ \cline{3-14}
\multicolumn{2}{c|}{}&MC    &VAR   &CTMC  &LR    &HP    &MPP   &SCP   &DMCP &SSCP &WDMCP &HDMCP &SDMCP\\ \hline
\multirow{8}{*}{$\mbox{AC}_d$'s}&1-day  &\textbf{1.000}&\textbf{1.000}&0.017&0.603&0.445&0.656&0.618&0.652
&0.628&0.646&0.677&0.705\\
&2-day  &0    &0.002&0    &0.481&0.385&0.509&0.568&0.633&0.605&0.599&0.329&\textbf{0.683}\\
&3-day  &0    &0    &0    &0.115&0.350&0.158&0.135&\textbf{0.538}&0.178&0.185&0.187&0.297\\
&4-day  &0    &0    &0.215&0.134&0.150&0.165&0.158&0.105&0.285&0.204&0.256&\textbf{0.321}\\
&5-day  &0    &0    &0.193&0.004&0.029&0.035&0.035&0.052&0.240&0.106&0.154&\textbf{0.305}\\
&6-day  &0    &0    &\textbf{0.557}&0.008&0.058&0.019&0.022&0.054&0.238&0.097&0.162&0.335\\
&7-day  &0    &0    &0    &0.003&0    &0.009&0.016&0.029&0.237&0.059&0.109&\textbf{0.340}\\
&$>$7-day  &0    &0.005&0    &0.560&0    &0.574&0.378&\textbf{0.734}&0.357&0.676&0.652&0.715\\ \hline
$\mbox{AC}_D$&Overall&0.171&0.173&0.058&0.355&0.253&0.387&0.357&0.508&0.412&0.443&0.389&\textbf{0.542}\\
        \hline\hline
      \end{tabular}
   \end{threeparttable}
\end{table*}

\begin{table*}[t!]
  \centering
  \small\caption{Overall prediction accuracy for various methods on relative simulation errors}\label{tab:error}
  \begin{threeparttable}[c]
      \begin{tabular}{
        @{\hspace{6pt}}c|c|c@{\hspace{6pt}}
        @{\hspace{6pt}}c@{\hspace{6pt}}
        @{\hspace{6pt}}c@{\hspace{6pt}}
        @{\hspace{6pt}}c@{\hspace{6pt}}
        @{\hspace{6pt}}c@{\hspace{6pt}}
        @{\hspace{6pt}}c@{\hspace{6pt}}
        @{\hspace{6pt}}c@{\hspace{6pt}}
        @{\hspace{6pt}}c@{\hspace{6pt}}
        @{\hspace{6pt}}|c@{\hspace{6pt}}
        @{\hspace{6pt}}c@{\hspace{6pt}}
        @{\hspace{6pt}}c@{\hspace{6pt}}
        @{\hspace{6pt}}c@{\hspace{6pt}}
        }
        \hline\hline
       \multicolumn{2}{c|}{\multirow{2}{*}{Method}}&\multicolumn{8}{c}{Without pre-processing} & \multicolumn{4}{|c}{With pre-processing}\\ \cline{3-14}
\multicolumn{2}{c|}{}&MC    &VAR   &CTMC  &LR    &HP    &MPP   &SCP   &DMCP &SSCP &WDMCP &HDMCP &SDMCP\\ \hline
\multirow{8}{*}{$\mbox{Err}_c$'s}&CCU     &0.799&0.803&0.942&0.256&0.247&0.230&0.453&0.230&0.398&0.316&0.290&\textbf{0.201}\\
&ACU     &1.002&0.903&1.003&0.838&0.855&0.790&0.903&0.468&0.894&0.611&0.433&\textbf{0.406}\\
&FICU    &0.864&0.861&0.664&0.311&0.329&0.287&0.553&0.281&0.508&0.357&0.288&\textbf{0.245}\\
&CSRU    &0.504&0.506&1.042&0.148&0.156&0.172&0.295&0.142&0.224&0.206&0.168&\textbf{0.131}\\
&MICU    &0.821&0.819&0.877&0.196&\textbf{0.191}&0.231&0.442&0.223&0.386&0.308&0.259&0.197\\
&TSICU   &0.767&0.760&0.933&0.407&0.393&0.420&0.580&0.273&0.538&0.386&0.398&\textbf{0.259}\\
&NICU    &0.903&0.903&0.373&0.064&\textbf{0.059}&0.068&0.227&0.114&0.150&0.149&0.141&0.100\\
&GW      &1.536&1.535&0.908&0.294&0.296&0.293&0.464&0.230&0.610&0.323&0.277&\textbf{0.208}\\ \hline
$\mbox{Err}_C$&All CUs&0.984&0.982&0.730&0.215&0.218&0.224&0.419&0.204&0.395&0.281&0.243&\textbf{0.181}\\
        \hline\hline
      \end{tabular}
   \end{threeparttable}
\end{table*}

\subsection{Comparison Results}
We compare our DMCP method with other competitors on predicting destination CUs and duration days in current CUs, and simulating the dynamics of patient flow. 
The prediction results are shown in Fig.~\ref{FigAcc}, and the relative simulation errors are shown in Fig.~\ref{FigSim}. 
The numerical results of overall prediction accuracy and simulation error are shown in Tables~\ref{tab:acc},~\ref{tab:acd}, and~\ref{tab:error}. 
Experimental results of these three tasks show that our DMCP method obtains superior results in most situations and outperforms other methods. 
Furthermore, adding proposed data synthesis method as the pre-processing of training data, our SDMCP method further improves the testing results. 
Specifically, we can find that:

1) According to Fig.~\ref{FigAcc}, Fig.~\ref{FigSim}, and Tables~\ref{tab:acc},~\ref{tab:acd}, we can find that our DMCP methods obtain the highest overall prediction accuracy and the lowest simulation error. 
Compared with the second best methods, i.e., the HP for predicting destination CUs and the MPP for predicting duration days, our DMCP achieves improvements over 4\% and 11\% respectively. 
The encouraging results demonstrate that our mutually-correcting process model is suitable for describing patient flow. 

2) The feature-independent methods (MC, VAR and CTMC) perform poorly in all three tasks. 
Because of the imbalance of data, there are insufficient transition processes involving those rarely-used CUs. 
For these CUs, the transition probabilities learned via MC and CTMC and the transition coefficients learned via VAR are unreliable. 
For example, in Fig.~\ref{FigAccC}, we can find that these methods only obtain high accuracy for general ward because it is contained via most patients' transition processes. 
For other CUs, however, the prediction accuracy is almost zero in most situations. 
Similar phenomenon can also be observed in the prediction results of duration days --- only the 1-day situation is predicted with high accuracy while the rest situations cannot be predicted. 

3) Compared with feature-independent methods, the LR method improves the testing results greatly, which demonstrates the importance of EHR-based features for predicting patient flow. 
Applying EHR-based features suppresses the negative influence caused by imbalanced data and improves the prediction results of the classes having insufficient samples. 
Specifically, in Fig.~\ref{FigAcc} we can find that LR outperforms MC, VAR, and CTMC in most situations, whose overall accuracy is improved over 20\% in both prediction tasks. 

4) The point process-based methods (HP, MPP, SCP, and our DMCP) further improve the prediction accuracy for both two learning tasks because of considering the temporal influences of historical features on current predictions. 
Specifically, the HP method trains a Hawkes process model in a generative way, and the joint probability $p(c,d,t|\mathcal{H}_t^u)$ is estimated. 
However, as aforementioned, such a generative learning method is sensitive to the insufficiency and imbalance of data. 
As a result, the predictive model does not work when it comes to predict the classes having few samples, i.e., ACU, FICU, and TSICU in Fig.~\ref{FigAccC}, and the duration with 7-day in Fig.~\ref{FigAccD}. 
On the contrary, the discriminative learning methods (MPP, SCP and our DMCP) are more robust, which improves prediction results in most situations, especially the classes having few samples. 

5) Adding suitable pre-processing in the training phase indeed enhances the robustness of our DMCP method to imbalanced data and improves the testing results. 
In Tables~\ref{tab:acc},~\ref{tab:acd}, and~\ref{tab:error}, we can find that because of the weaknesses analyzed in Section~\ref{sec:ub}, WDMCP and HDMCP are slightly inferior to original DMCP method. 
Fig.~\ref{FigAcc} illustrates the reason obviously: while the prediction accuracy for those minor classes, i.e., the ACU, FICU in Fig.~\ref{FigAccC} and the $4$-day in Fig.~\ref{FigAccD}, is improved, the performance on major classes degrades more, i.e., the CSRU, NICU in Fig.~\ref{FigAccC} and the $2$-and $3$-day in Fig.~\ref{FigAccD}. 
For WDMCP, increasing the weight of some training samples, especially for some minority sample classes with only a few samples, may lead to over-fitting of the classifier, so the performance is bad when it comes to the testing set that may only have a slight difference from the training set. 
For HDMCP, the performance is degraded because the linear-inseparable property of MINORITY class in each step increases the difficulty of training. 
The proposed SDCMP, on the contrary, improves the result of minor classes and avoids the degradation of the result of major classes jointly, which obtains even better results than original DMCP --- both the $\mbox{AC}_C$ and $\mbox{AC}_D$ increase over $3$\% and the $\mbox{Err}_C$ is reduced to $0.183$. 

Additionally, all the methods above are stable with the change of training data. 
In the case of 
Using 10-fold cross validation, the fluctuations of their testing results are all within $\pm 0.01$.

\subsection{Feature Selection Result}
As aforementioned, our method achieves feature selection via group lasso. 
Treating each dimension of feature as a group, we measure the importance of each group via the amplitude of the coefficient associated with the group, denoted as $|\bm{\Theta}_m |$. 
The large amplitude means that the change of feature corresponding to the coefficient has a large influence on the prediction result.
Specifically, when the coefficient is zero, it means that the corresponding feature does not change the conditional intensity function, and therefore, has no influence on the transition to the destination CU and the duration time. 
When the coefficient is positive, it means that the corresponding feature will increase the conditional intensity function. 
Such a feature (profile, treatment, nursing operation, or medication) increases the probability that transiting patients to certain CUs and staying certain days. 
On the contrary, when the coefficient is negative, the corresponding feature decreases the probability of certain transition events.

\begin{figure*}[!t]
\centering
\subfigure[The distribution of coefficients for predicting destination CUs.]{
\includegraphics[width=1\linewidth]{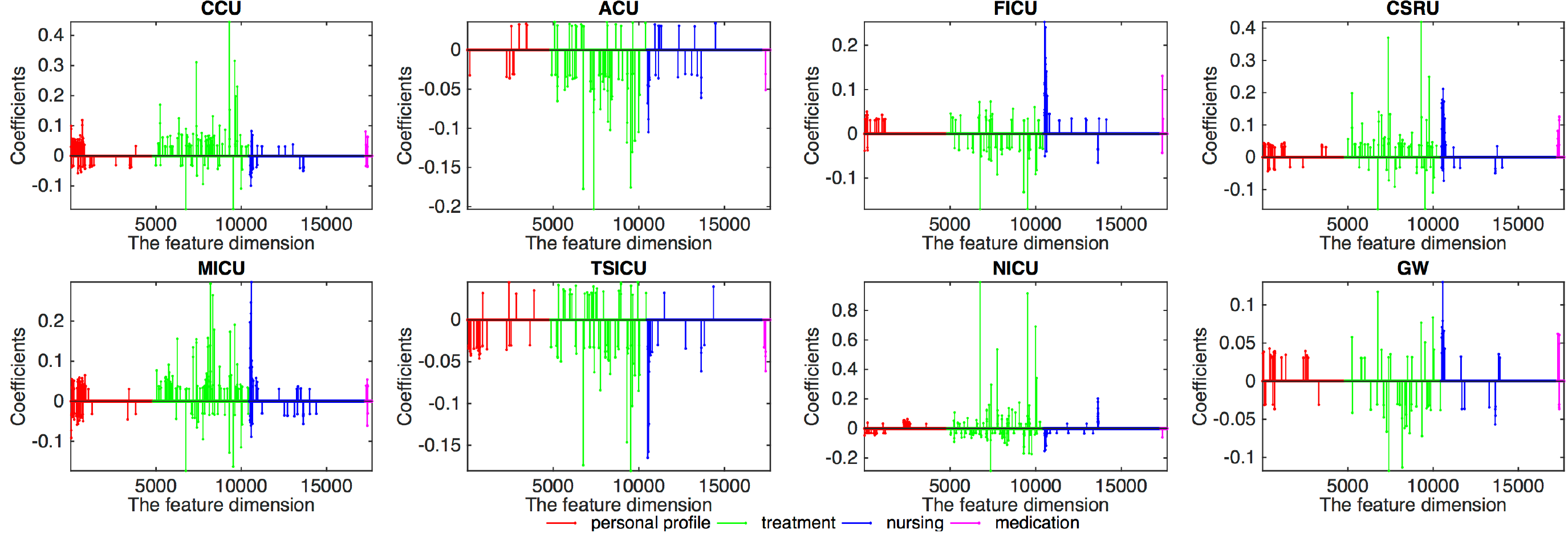}\label{Fig8a}
}
\subfigure[The distribution of coefficients for predicting duration days.]{
\includegraphics[width=1\linewidth]{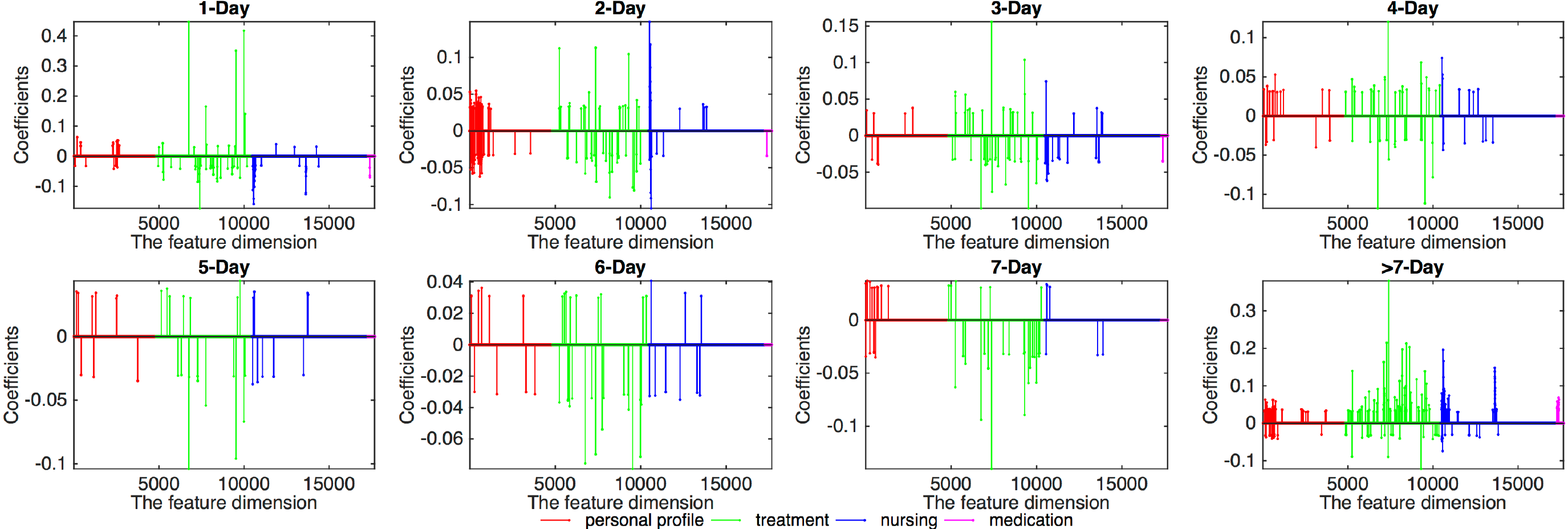}\label{Fig8b}
}
\caption{
Feature selection result.}%\label{Fig8}
\end{figure*}

Figs.~\ref{Fig8a} and~\ref{Fig8b} visualize the coefficients in different feature domains w.r.t. various learning tasks. 
We can find that most of the time-varying features related to treatments are selected via at least one learning task while the time-invariant features (personal profile) and the time-varying features related to nursing programs and medications focused on certain parts. 
Another interesting observation is that many features have negative coefficients. 
It means that these features suppress the transitions among CUs and lengthen the duration in current CU. 
We think these phenomena are reasonable based on the following reasons.
1) The treatments are the most influential factors for the patient flow, whose progresses and feedbacks impact on the transitions between CUs and the durations in them greatly. 
Therefore, it is natural that most of features in this domain are with large parameters.
2) The features across different dimensions in the personal profile domain are likely to be correlated with each other, i.e., a certain disease's diagnose is correlated with patient's age and gender. 
Therefore, only a part of features in this domain are selected.
3) Similarly, nursing programs and medications are highly correlated with the treatments. 
When most of features related to treatments are selected, only a part of them are useful.
4) Some diseases and corresponding treatments require patients to stay at certain CUs for a long time. 
When the treatments, nursing operations, or medications happen, the patients are unlikely to transit to other CUs in few days. 

\subsection{Impacts of Parameters}
The parameters in our method are the learning rate of gradient descent $\beta$, the bandwidth of Gaussian kernel $\sigma$ in our mutually-correcting process model, the weight of group-lasso $\gamma$, and the weight of augmented Lagrangian $\rho$.  
The learning rate $\beta$ controls the step length of gradient descent. 
Too large $\beta$ will lead our algorithm to be unstable while too small $\beta$ will lead our algorithm to converge too slowly.  
Following the work in~\cite{schaul2013no}, we set the learning rate $\beta$ decays with rate $\mathcal{O}(k^{-1})$, where $k$ is the number of iteration. 
Its initial value for our work is set as $10^{-4}$. 

The parameter $\sigma$ controls the importance of historical EHR-based features. 
When $\sigma$ is large, the kernel $\exp(-\frac{(t-t')^2}{\sigma^2})$ decays slowly, which means the temporal influence of historical events will exist for a long time. 
In an extreme case that $\sigma\rightarrow \infty$, the kernel will tend to be $1$, and our mutually-correcting process model will ignore the temporal difference among historical events and degrade to a self-correcting process. 
On the contrary, when $\sigma$ is small, the kernel decays rapidly and the influence of historical events will be short. 
In the case that $\sigma\rightarrow 0$, our model will only consider the feature at current time and our learning algorithm will be similar to the LR method mentioned above. 
For achieving a trade-off, we set $\sigma$ as the mean of duration days in our work. 

\begin{figure}[!t]
\centering
\subfigure[$\mbox{AC}_C$ and $\mbox{AC}_D$ w.r.t. $\gamma$.]{
\includegraphics[width=0.44\linewidth]{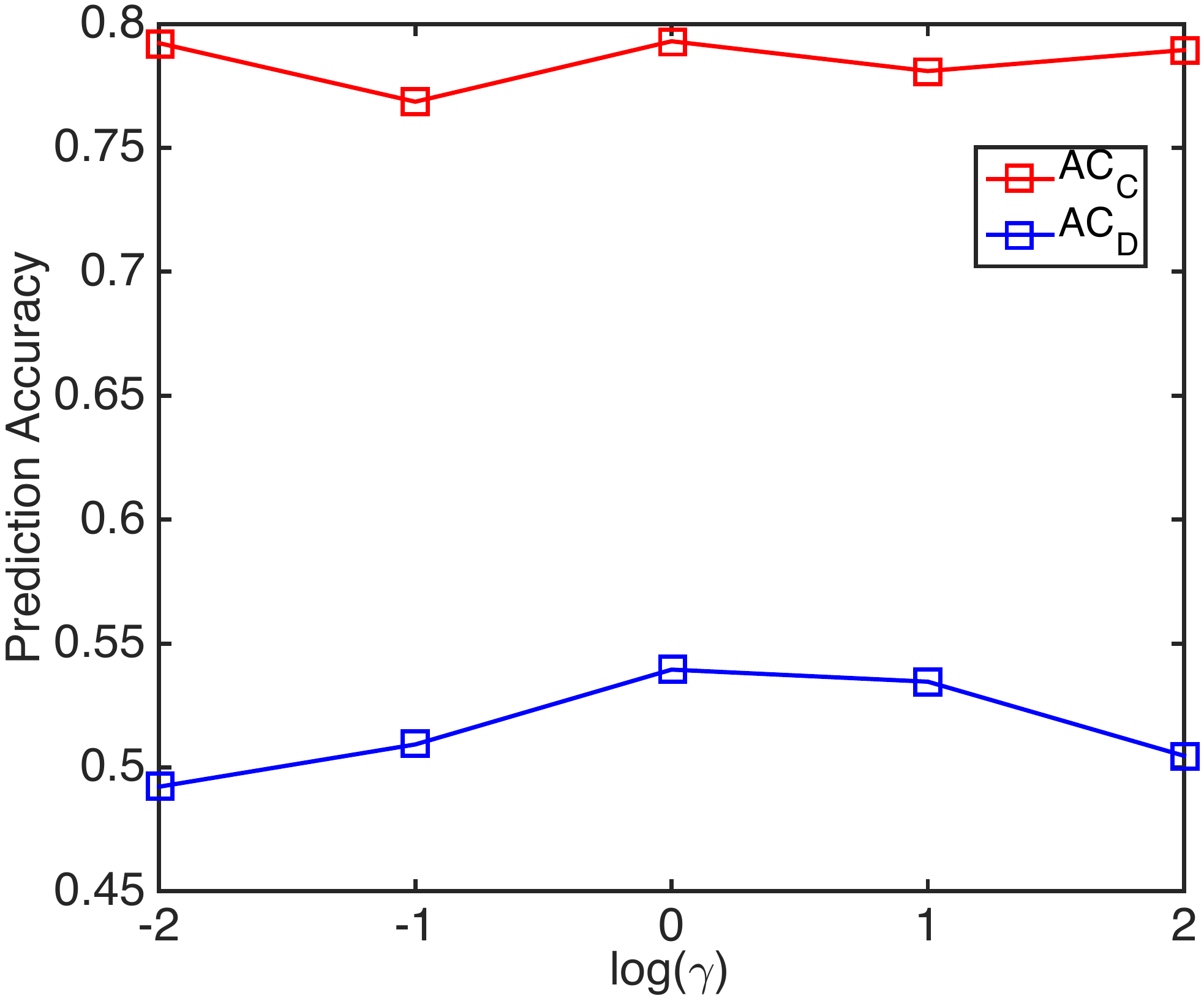}\label{Figgamma}
}
\subfigure[$\mbox{AC}_C$ and $\mbox{AC}_D$ w.r.t. $\rho$.]{
\includegraphics[width=0.44\linewidth]{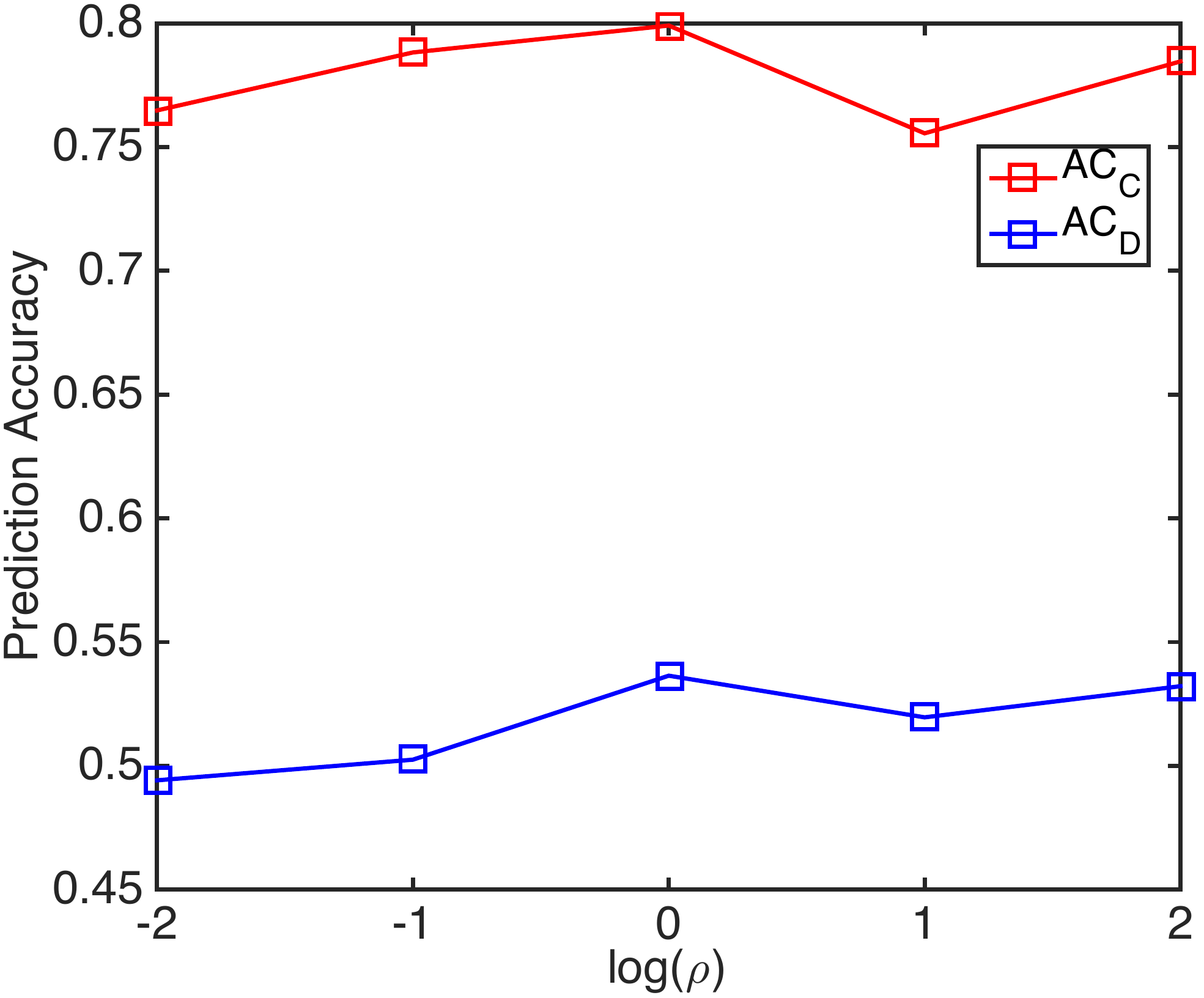}\label{Figrho}
}
\caption{
The overall AC of our method w.r.t. the changes of $\gamma$ and $\rho$. (a) When we investigate the robustness of our method to $\gamma$, we set $\rho=1$.  
(b) When we investigate the robustness of our method to $\gamma$, we set $\gamma=1$.}\label{FigGamma}
\end{figure}

We also investigate the robustness of our method to the changes of $\gamma$ and $\rho$.
The parameter $\gamma$ controls the importance of group-lasso. 
In the case that the features of data indeed yield to the assumption of group sparsity, a suitable $\gamma$ will regularize model well and improve the result of feature selection, while too large or too small $\rho$ will cause the misspecification of model. 
Fig.~\ref{Figgamma} gives the overall AC of our method w.r.t. the change of $\gamma$. 
We can find that the learning result is relatively stable in a wide range of $\gamma$ and the result corresponding to $\gamma=1$ is slightly better than others. 

The parameter $\rho$ reflects the importance of augmented Lagrangian. 
It mainly controls the convergence rate of ADMM algorithm~\cite{nishihara2015general}. 
Fig.~\ref{Figrho} gives the overall AC of our method w.r.t. the change of $\rho$. 
We can find that the learning result is very stable in a wide range of $\rho$. 
A slightly degradation happens when we set a large $\rho$. 
In such a situation, the step length in Eq.~(\ref{updateTheta}) will be too large and cause the oscillatory updating around the optimal point. 

In summary, our SDMCP method is robust to these two parameters. 
According to Fig.~\ref{FigGamma}, we set $\gamma=1$ and $\rho=1$ empirically. 

\section{Related Work}\label{sec2:relatedwork}
\subsection{EHR and Feature Representation} 
A typical electronic health record consists of a patient's profile (i.e., gender, age), her diagnose of certain diseases (i.e., ICD code), and her treatment process, e.g., medications, nursing information, the transitions and durations in various care units. 
An important application driven via EHRs is extracting characteristic features of physiology in clinical data, or called phenotyping~\cite{hripcsak2013next}.  
In~\cite{liu2015temporal}, the temporal phenotyping from EHRs is achieved by a graph-based model, where a temporal graph of patients' events (i.e., diagnoses and treatments of diseases) is constructed and phenotypes are extracted via decomposing the adjacent matrix of the graph. 
In~\cite{wang2015rubik}, a binary tensor indicating patients' diagnose and the medications they used is given and phenotypes are extracted via non-negative factorization of the tensor with sparse constraints. 
In~\cite{che2015deep}, the deep computational phenotyping is achieved via stacked autoencoder. 
All these works can be viewed as feature extraction methods for EHRs. 
The feature obtained via these works can be further applied to other problems like constructing disease network~\cite{choi2015constructing} and modeling patient flow~\cite{ozkaynak2015characterizing} that we care in this paper.

\subsection{Patient Flow and Traditional Models}
Many patient flow models based on EHRs have been proposed for recent years. 
The early work in~\cite{ceglowski2005facilitating} models patient flows from a viewpoint of treatment processes and proves that the treatment clustering information helps to model patient flow in emergency departments indeed. 
Following this strategy, the information of patients' treatment types is used to estimate the crowdedness of emergency departments in~\cite{ceglowski2006investigation}. 
For example, the workflow of emergency departments is modeled based on the features extracted from patients' EHRs in~\cite{ozkaynak2013patient} and the work is further specialized for pediatric asthma patients in~\cite{ozkaynak2015characterizing}. 
Additionally, the visualization and analysis of patient flow are achieved jointly in~\cite{wongsuphasawat2011outflow,vankipuram2011toward} based on patients' EHRs.
Most of methods above are based on EHRs formulated as time series. 
Many traditional models, such as Markov chain (MC) model~\cite{cole2005multistate}, vector auto-regressive (VAR) model~\cite{arnold2007temporal,meek2014toward,han2013transition} and hidden Markov model (HMM)~\cite{rabiner1986introduction,cooper2004analysis,ginter2009combining}, can be used to model patients' transition processes among different states. 
However, the works above mainly focus on modeling the flow of patients having a certain kind of diseases from discrete time series or aggregate data. 
None of them attempt to model general patient flow in continuous time.

\subsection{Continuous-time Models}
Recently, many efforts have been made to extend the models above from discrete time domain to continuous one. 
The continuous-time Markov chain (CTMC) is proposed in~\cite{iannelli2014continuous} to model the Markov chain in continuous time domain, which can be viewed as a special case of semi-Markov models~\cite{krol2015semimarkov}. 
Similarly, a hidden Markov model in continuous time domain is proposed in~\cite{liu2015efficient}. 
Focusing on e-health related applications, these continuous-time models have been widely used to analyze EHRs.
For example, in~\cite{zhaomining,choi2015constructing}, Hawkes process-based models are proposed to capture the temporal triggering patterns between diseases. 
A continuous-time HMM is proposed in~\cite{liu2015efficient} to model the progression of diseases.

Point processes are a kind of classic tools for modeling continuous-time event sequences~\cite{daley2007introduction}. 
Many different point processes have been proposed for various applications, e.g., the Hawkes processes for social network modeling~\cite{yang2013mixture,li2014learning,zhao2015seismic} and information system analysis~\cite{luo2015multi,yan2015machine}, and the self-correcting processes for earthquake prediction~\cite{isham1979self,ogata1984inference} and vision perception model~\cite{xu2015trailer}. 
An advantage of these point process models is considering the influence of all historical events on current one, which make these models outperform traditional low-order Markovian models.
Recently, some works start to apply point process-based model to analyze EHRs for health information systems~\cite{lian2015multitask,zhaomining}. 

As aforementioned, it is surprising that very few works make attempts to model and predict patient flow via a continuous-time model. 
Additionally, from the viewpoint of methodology, all the methods above are generative.
The joint distribution of all transitions in the continuous domain are learned via the maximum likelihood estimator. 
However, because of the following two reasons, sometimes it is necessary for us to propose a discriminative model. 
One reason is in some learning tasks, e.g., predicting future transitions, we care more about the conditional probability of current transition given historical transitions rather than the joint probability of all transition events. 
The other is facing sparse or imbalanced data, learning a generative model may suffer to serious over-fitting problem. 
Unfortunately, the discriminative learning methods for continuous-time models like point processes are not explored in depth.

\subsection{Imbalanced Data Processing}
Many methods have been proposed to learn models from imbalanced data. 
Generally, these methods can be categorized into two classes. 
One kind of the methods is merging minor classes together and learning binary classifiers step-by-step~\cite{sethi1982hierarchical,longford1987fast}. 
Another is weighting training samples to re-balance data~\cite{king2001logistic,tan2005neighbor,zhuang2005efficient}, where the samples in the minor classes have large weights while those in the major ones have small weights. 
This kind of methods are extended recently in~\cite{lee2003learning}.
The weights are added to unlabeled samples when training logistic regression, which can be viewed as the prior knowledge of model. 
More recently, the imbalanced data processing methods based on auxiliary samples are proposed. 
In~\cite{xu2015dictionary}, a classifier based on semi-supervised dictionary learning is proposed for the classes with extremely few samples. 
Unlabeled samples are used as auxiliary samples in the training phase and added to minor classes adaptively. 
Focusing on the problem of data synthesis, auxiliary data is generated based on the manifold learning in~\cite{Xu_2013_ICCV,xu2014manifold}.
However, they do not consider the data imbalance problem in the classification task.

\section{Conclusion}\label{sec:con}
Focusing on predicting patient flow, we propose a novel mutually-correcting process model and its discriminative learning algorithm in this paper. 
Our mutually-correcting process model improves the flexibility of existing parametric point process models, which reflects the properties of patient flow. 
The proposed discriminative learning algorithm combines multinomial logistic regression with group-lasso, and achieves feature selection during learning model. 
We also consider the data imbalance problem in the real-world dataset and propose a novel pre-processing method for training samples, which greatly improves the learning result. 
Compared with the state-of-art methods, our method obtains superior prediction results on real-world data set, which has potential to predict overcrowdedness or conflicted usage of CUs in practical situations. 
Our method is applicable to modeling a patient's need for various ``care teams'' within the CU (critical care nurses, a pharmacist, a nutritionist, respiratory therapists, consultants, social workers and case managers, clergy, etc), which will further improve care management and coordination for patients with multiple chronic conditions. 

%\textcolor{black}{
It should be noted that the proposed work is a first step towards our goal that predicting and managing patient flow. 
Many problems are not completely solved, which will be our future work. 
For example, although our method is superior to other competitors in most situations, we can find that for the transitions and the duration days happening with low frequency, the prediction results obtained by our method are still unsatisfying. 
It means that the robustness problem to imbalanced data is still not completely solved, which is one direction of our future research work. 
Another problem is the prediction accuracy of duration time. 
Currently, we can merely predict the duration time accurate to ``day'', which is too coarse for practical situations. 
In the future, we will make efforts to extend our methodology and further improve the prediction accuracy of duration time.%}
Additionally, we also plan to extend our mutually-correcting process to a nonparametric model.

\section{Acknowledgment}
This work is supported in part by NIH/NSF BIGDATA R01 GM108341, NSF IIS-1639792, NSF DMS-1620345, and NIH early career development award in biomedical big data science (1K01ES025445-01A1).

\bibliographystyle{IEEEtran}
\bibliography{sigproc}

\end{document}